\documentclass[10pt,journal,compsoc]{IEEEtran}
\usepackage[utf8]{inputenc}
\usepackage{tikz}
\usepackage{comment}
\usepackage{amsmath,amssymb} 
\usepackage{color}
\usepackage{wrapfig}
\usepackage{rotating}
\usepackage{epsfig}
\usepackage{graphicx}
\usepackage{multirow}
\usepackage{subcaption}
\usepackage{tabularx}
\usepackage{floatrow}

\usepackage[pagebackref,breaklinks,colorlinks]{hyperref}

\ifCLASSOPTIONcompsoc
  \usepackage[nocompress]{cite}
\else
  \usepackage{cite}
\fi

\begin{document}

\title{Image-to-Image Translation with  Disentangled Latent Vectors for Face  Editing}

\author{
Yusuf Dalva, Hamza Pehlivan, Oyku Irmak Hatipoglu, Cansu Moran, and Aysegul Dundar
\IEEEcompsocitemizethanks{
\IEEEcompsocthanksitem Y. Dalva is with the Department of Computer Science, Virginia Tech, USA.
\IEEEcompsocthanksitem H. Pehlivan is with MPI, Germany.
\IEEEcompsocthanksitem O. I. Hatipoglu is with ETH, Zurich.
\IEEEcompsocthanksitem C. Moran is with TUM, Germany.
\IEEEcompsocthanksitem A. Dundar is with the Department of Computer Science, Bilkent  University,
Ankara, Turkey.%
\IEEEcompsocthanksitem Corresponding author: A. Dundar, Email: adundar@cs.bilkent.edu.tr 
}
}

\IEEEtitleabstractindextext{
\begin{abstract}
We propose an image-to-image translation framework for facial attribute editing with disentangled interpretable latent directions.
Facial attribute editing task faces the challenges of targeted attribute editing with controllable strength and  disentanglement in the representations of attributes to preserve the other attributes during edits.
For this goal, inspired by the latent space factorization works of fixed pretrained GANs,  we design the attribute editing by latent space factorization, and for each attribute, we learn a linear direction that is orthogonal to the others.
We train these directions with orthogonality constraints and disentanglement losses.
To project images to semantically organized latent spaces, 
we set an encoder-decoder architecture with attention-based skip connections.
We extensively compare with previous image translation algorithms and editing with pretrained GAN works. 
Our extensive experiments show that our method significantly improves over the state-of-the-arts. 
Project page: \href{https://yusufdalva.github.io/vecgan}{https://yusufdalva.github.io/vecgan}
\end{abstract}

}

\maketitle

\IEEEdisplaynontitleabstractindextext
\IEEEpeerreviewmaketitle

\section{Introduction}
\newcommand{\interpfigz}[1]{\includegraphics[trim=0 0 0cm 0, clip, width=3.2cm]{#1}}
\begin{figure*}
\centering
\scalebox{0.71}{
\addtolength{\tabcolsep}{-5pt}   
\begin{tabular}{cccccc}
\\
\interpfigz{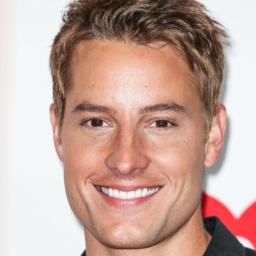} &
\interpfigz{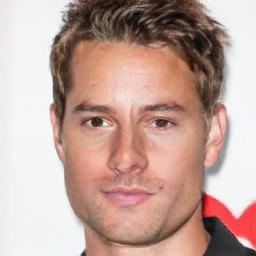} &
\interpfigz{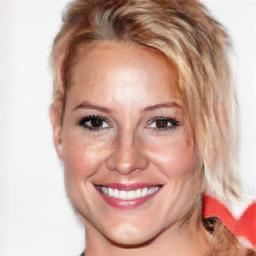} &
\interpfigz{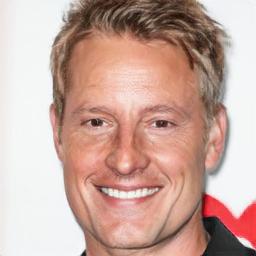} &
\interpfigz{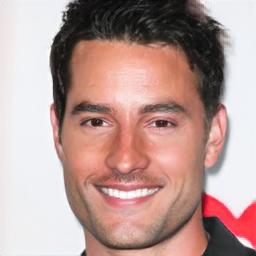} &
\interpfigz{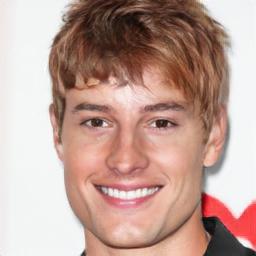} 
\\
\interpfigz{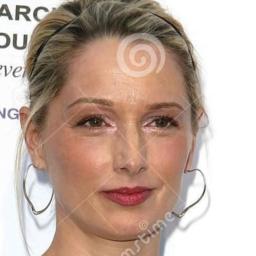} &
\interpfigz{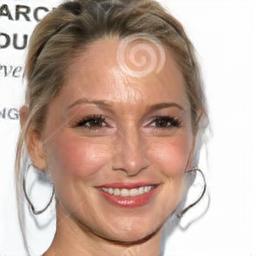} &
\interpfigz{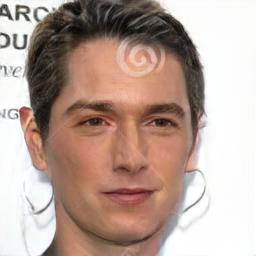} &
\interpfigz{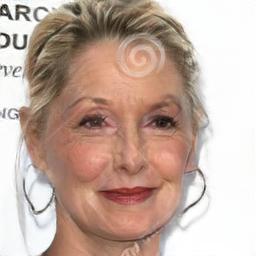} &
\interpfigz{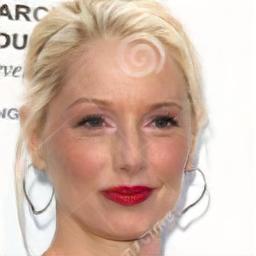} &
\interpfigz{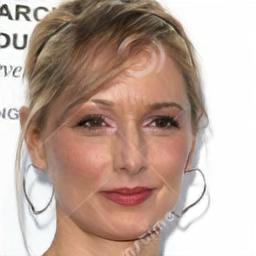} 
\\
Input &  Smile &  Gender & Age  & Hair Color &  Bangs \\
\end{tabular}
}
\caption{VecGAN++ image translation results.}
\label{fig:results_teaser}
\end{figure*}

Facial attribute editing task has been a popular topic among the image translation tasks, and significant improvements have been achieved with generative adversarial networks (GANs) \cite{starganv2,shen2017learning,xiao2018elegant,zhang2018generative,li2021image, dalva2022vecgan}.
Facial attribute editing is one of the most challenging translation tasks because only one attribute of a face is expected to be modified without affecting other attributes, whereas humans are very good at detecting if a person's identity or any other attributes of the face change.
There are two main directions proposed for facial attribute editing, 1- end-to-end trained image translation networks and 2- latent manipulation of pretrained GAN networks.
For the first one, different image translation architectures are proposed with usually two networks, one for encoding style and the other for image editing with modified styles are injected into \cite{starganv2,li2021image}.
A style, an attribute, is usually encoded from another image or  sampled from a distribution.
The attributes both in the encoding and editing phases need to be disentangled.
To achieve such disentanglement, works focus on style encoding and progress from a shared style code, SDIT \cite{wang2019sdit}, to mixed style codes, StarGANv2 \cite{starganv2}, to hierarchical disentangled styles, HiSD \cite{li2021image}.
Among these works, HiSD independently learns styles of each attribute, bangs, hair color, and eyeglasses, and introduces a local translator which uses attention masks to avoid global manipulations.
HiSD showcases successes on those three local attribute editing tasks and is not tested for global attribute editing, e.g., age, smile.
Furthermore, one limitation of these works is the uninterruptible style codes, as one cannot control the intensity of attributes (e.g., blondness) in a straightforward manner.

The second class of methods builds on well-trained generative models, specifically StyleGAN2 models \cite{karras2020analyzing}, which organize their latent space as disentangled  representations with meaningful directions in a completely unsupervised way.
This approach has two steps; in the first step, an input image is embedded into the generative model's latent space via additional training of an encoder or latent optimization. In the second step, the embedded latent code is modified based on the discovered directions such that when the edited code is decoded, an attribute is edited in the input image. 
Embedding images in GAN's space and exploring interpretable directions in latent codes have emerged as important research endeavors on the fixed pretrained GANs \cite{shen2020interpreting,voynov2020unsupervised,harkonen2020ganspace}.
We refer to these models as  \textit{StyleGAN inversion-based methods} throughout the manuscript.
However, these models are not trained end-to-end. Therefore, there is no guarantee that the encoded image lies in the natural GAN space, which results in the codes with limited editability. When encoders are forced to encode images into GAN's natural latent distribution, the results suffer from the lack of faithful reconstruction.
Additionally, since the model is not trained for this task, there is no guarantee that interested attributes would be disentangled (e.g. eyeglasses - age).

To overcome the challenges of facial attribute editing task, our previous work proposed an image-to-image translation framework with interpretable latent directions, VecGAN \cite{dalva2022vecgan}. 
The attribute editing directions of VecGAN are learned in the latent space and regularized to be orthogonal to each other for style disentanglement.
The attribute edits are achieved by linear operations in the latent space with a deep encoder-decoder as explained in details in Sec. \ref{sec:generator} and \ref{style-shift}.
In this work, we improve upon VecGAN \cite{dalva2022vecgan} with a novel disentanglement loss and architectural changes.
This manuscript extends its conference version \cite{dalva2022vecgan}  with the following additions:
\begin{itemize}
    \item We propose disentanglement loss in Sec. \ref{sec:objectives} that stabilizes the training and improves the results.
    \item  We augment our generator with attention-based skip connections which is encoded from the feature maps of original and edited features. This way, the network can pass only selected information to the decoder. The module is explained in Sec.
    \ref{sec:skip}. 
    We refer to our final  model with updated architecture and loss objective as VecGAN++. Results are shown in Fig. \ref{fig:results_teaser}.
    \item We compare our method with several state-of-the-art image translation methods, especially with popular pretrained GAN-based models. We provide results with an extensive number of metrics for quality, attribute edit accuracy, identity, and background preservation (Sec. \ref{sec:metrics}). Our results show the effectiveness of our framework with significant improvements over state-of-the-art as provided in Sec. \ref{sec:results}.
    \item We provide a comprehensive analysis of our results in Sec. \ref{sec:analysis}. We report metrics as the strength of editing increases for our and competing methods. We also analyze the projected style codes and show that they can classify the targeted attributes of images, e.g. hair color, smile.
\end{itemize}
\section{Related Work}

\textbf{Image to Image Translation.} These algorithms are trained end-to-end to change selected attributes of images while preserving the content\cite{dundar2020panoptic, liu2022partial, starganv2,shen2017learning,xiao2018elegant,zhang2018generative,li2021image,hou2022guidedstyle, lyu2021sogan}.
They set an encoder-decoder architecture and train the models with reconstruction and GAN losses.
For the multi-modal image translation models, style is encoded separately from another image or sampled from a distribution \cite{huang2018multimodal,starganv2}.
The generator, decoder, receives style and content information separately
 \cite{zhu2017multimodal, li2021image, zhu2020sean}.
 These works use two encoders, one is to extract style, and one is to encode content \cite{li2021image,yang2021l2m, lyu2023dran}.
 However, in many cases, it is not clear what the content is and what the style is. Usually, style is referred to the domain attributes one wants to change, and the content is the rest of the attributes, which is an ambiguous problem definition.
In our work, we design the attribute as disentangled directions in the latent space, and we do not employ separate style and content encoders.

\noindent \textbf{Editing with pretrained GANs (StyleGAN inversion-based models).} Facial attribute editing is also shown to be possible with pretrained GANs. State-of-the-art GAN models organize their latent space with interpretable directions \cite{karras2019style,karras2020analyzing,yu2021dual,yildirim2023diverse} such that moving along the direction only changes one attribute of the image\cite{shen2020interpreting, voynov2020unsupervised,harkonen2020ganspace,shen2021closed}, and many directions are found for face editing, e.g. directions that change the smile, pose, age attributes are found, to name a few.
To edit a facial attribute of an input image, one needs to project the image to a latent code in GANs' latent space such that the generator reconstructs the input image from the latent code \cite{abdal2019image2stylegan, abdal2020image2stylegan++}.
There have been various architectures \cite{alaluf2022hyperstyle, wang2022high} and objectives proposed to project an image to GAN's embedding. However, they suffer from reconstruction-editability trade-off \cite{tov2021designing}.
That is if the image is faithfully reconstructed, it may not lie in the true distribution of GANs latent space, and therefore, the directions do not work as expected, which prevents editing the image.
On the other hand, if the projection is close to the true distribution, then the reconstruction is poor.
We also show this behavior in the Results section \ref{sec:results} when comparing our method with state-of-the-art editing with pretrained GANs methods.
On the other hand, we are inspired by the natural organization of well-trained GANs and our image-to-image translation framework is designed similarly with disentangled latent space manipulations. 

\section{Method}

In this section, we provide an overview of the generator architecture and the training set-up.
We follow the hierarchical labels defined by \cite{li2021image}.
For a single image, its attribute for tag $i \in \{1,2,...,N\}$ can be defined as $j \in \{1,2,...,M_i\}$, where N is the number of tags and $M_i$ is the number of attributes for tag $i$.
For example, $i$ can be the tag of hair color, and attribute $j$ can take the value of black, brown, or blonde.

Our framework has two main objectives. As the main task, we aim to be able to perform the image-to-image translation task in a feature (tag) specific manner. While performing this translation, as the second objective, we also want to obtain an interpretable feature space that allows us to perform tag-specific feature interpolation.

\begin{figure*}[t]
    \centering
    \includegraphics[width=1.0\textwidth]{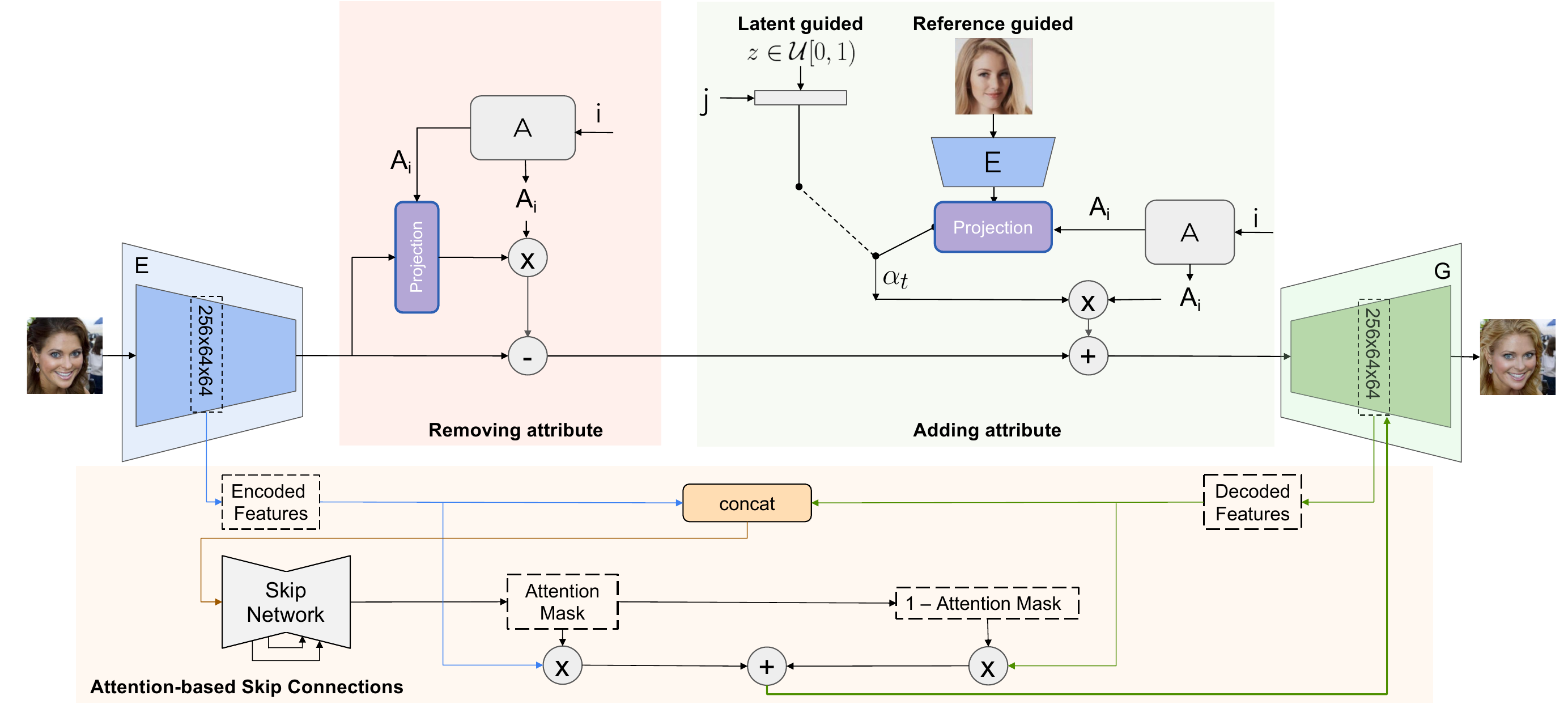}
    \caption{Our translator is built on the idea of interpretable latent directions. We encode images with an Encoder to a latent representation from which we change a selected tag ($i$), e.g. hair color with a learnable direction $A_i$ and a scale $\alpha$. To calculate the scale, we subtract the target style scale from the source style.
    This operation corresponds to removing an attribute and adding an attribute.
    To remove the image's attribute, the source style is encoded and projected from the source image.
    To add the target attribute, the target style scale is sampled from a distribution mapped for the given attribute ($j$), e.g. black, blonde, or encoded and projected from a reference image. 
    We also propose an attention-based skip connection module to transfer selected features without an information bottleneck to the decoder.}
    \label{fig:generator}
\end{figure*}

\subsection{Generator Architecture}
\label{sec:generator}
For the image-to-image translation task, we set an encoder-decoder based architecture and latent space translation in the middle as given in Fig. \ref{fig:generator}.
We perform the translation in the encoded latent space, $e$, which is obtained by $e = E(x)$ where $E$ refers to the encoder. The encoded features go through a transformation $T$, which is discussed in the next section.
The transformed features are then decoded by $G$ to reconstruct the translated images.
The image generation pipeline following feature encoding is described in Eq. \ref{eqn:decoder}.
 \begin{align}
     e' = T(e, \alpha, i) \nonumber \\
     x' = G(e') \label{eqn:decoder}
 \end{align}
 
Previous image-to-image translation networks \cite{li2021image,yang2021l2m,starganv2} set a shallow encoder-decoder architecture to translate an image while preserving the content and a separate deep network for style encoding.
In most cases, the style encoder includes separate branches for each tag.
The shallow architecture used to translate images prevents the model from making drastic changes in the images, which helps preserving the person's identity.
Our framework is different as we do not employ a separate style encoder and instead have a deep encoder-decoder architecture for translation.
That is because to be able to organize the latent space in an interpretable way, our framework requires a full understanding of the image and, therefore, a larger receptive field which results in a deeper network architecture.
A deep architecture with decreasing size of feature size, on the other hand, faces the challenges of reconstructing all the fine details from the input image.

With the motivation of helping the network to preserve tag-independent features such as the fine details from the background, we use attention-based skip connections between our encoder and decoder as described in Section \ref{sec:skip}.

The architectural details of the encoder and decoder are as follows: 
For the encoder, following a $1\times1$ convolution, we use 8 successive blocks that perform downsampling, which reduces feature map dimensions to 1x1. In our decoder, we have an architecture symmetric to the encoder, which is composed of 8 successive upsampling blocks. Except for the last downsampling block and the first upsampling block, we use instance normalization denoted as (+IN). The channels increase as \{32, 64, 128, 256, 512, 512, 512, 1024, 2048\} (for output resolution $256\times256$) in the encoder and decrease in a symmetric way in the decoder.
Each DownBlock and UpBlock has a residual block with $3\times3$ convolutional filters followed by a downsampling and upsampling layer, respectively. For downsampling, we use average pooling; for upsampling, we use nearest-neighbor.
We use the LeakyReLU activation layer (with slope 0.2) and instance normalization layer in each convolutional module.

\subsection{Translation Module} \label{style-shift}
To achieve a style transformation, we perform the tag-based feature manipulation in a linear fashion in the latent space.
First, we set a feature direction matrix $A$, which contains learnable feature directions for each tag.
In our formulation, $A_i$ denotes the learned feature direction for tag $i$.
The direction matrix $A$ is randomly initialized and learned during training.
$A$ matrix has the dimension of $6\times2048$ for $6$ attributes and encoder having the last convolution with channel size of $2048$.
Our translation module is formulated in Eq. \ref{eqn:translator}, which adds the desired shift on top of the encoded features $e$ similar to \cite{voynov2020unsupervised}.

\begin{equation}
T(e, \alpha, i) = e + \alpha \times A_i
\label{eqn:translator}
\end{equation}

We compute the shift as given in Eq \ref{eqn:alphat}.

\begin{equation}
\alpha = \alpha_t - \alpha_s
\label{eqn:alphat}
\end{equation}

Since the attributes are designed as linear steps in the learnable directions, we find the style shift by subtracting the source attribute scale from the target attribute scale.
This way, the same target attribute $\alpha_t$ can have the same impact on the translated images no matter what the attributes were of the original images.
For example, if our target scale corresponds to brown hair, the source scale can be coming from an image with blonde or black hair, but since we take a step for the difference of the scales, they can both be translated to an image with the same shade of brown hair.

There are two alternative pathways to extract the target shifting scale for feature (tag) $i$, $\alpha_t$.
The first pathway, named the latent-guided path, samples a $z \in \mathcal{U}[0,1)$ and applies a linear transformation $\alpha_t = w_{i,j} \cdot z + b_{i,j}$, where $\alpha_t$ denotes sampled shifting scale for tag $i$ and attribute $j$. We learn linear transformation parameters $w_{i,j}$ and $b_{i,j}$ in training time.  
Here tag $i$ can be hair color, and attribute $j$ can be blonde, brown, or black hair. We learn a different transformation module for each attribute, denoted  as $M_{i,j}(z)$.
Since we learn a single direction for every tag, e.g. hair color, this transformation module can put the initially sampled $z$'s into the correct scale in the linear line based on the target hair color attribute.
As the other alternative pathway, we encode the scalar value $\alpha_t$ in a reference-guided manner.
We extract $\alpha_t$ for tag $i$ from a provided reference image by first encoding it into the latent space, $e_r$, and projecting $e_r$ by $A_i$ as given in Eq. \ref{eqn:proj}.

\begin{equation}
\alpha_t = P(e_r, A_i) =  \dfrac{e_r \cdot A_i}{||A_i||}
\label{eqn:proj}
\end{equation}

In the reference guidance set-up, we do not use the information of attribute $j$, since it is encoded by the tag $i$ features of the image.

The source scale, $\alpha_s$, is obtained in the same way we obtain  $\alpha_t$ from the reference image.
We perform the projection for the corresponding tag we want to manipulate, $i$, by $P(e, A_i)$.
We formulate our framework with the intuition that the scale controls the amount of features to be added.
Therefore, especially when the attribute is copied over from a reference image, the amount of features that will be added will be different based on the source image.
For this reason, we find the amount of shift by subtraction as given in Eq. \ref{eqn:alphat}.
Our framework is intuitive and relies on a single encoder-decoder architecture.

\subsection{Attention-based Skip Connections}
\label{sec:skip}
We benefit from attention-based skip connections that merge encoded and decoded features with an attention mask to enable feature-aware residual connections. Our architecture includes a skip network $S$, which calculates an attention map from the concatenation of encoded and decoded features (feature resolution is 64x64 as illustrated in Fig. \ref{fig:generator}). The equation summarizing our approach is provided in Eq. \ref{eqn:attention} for encoded features $e$ and decoded features $d$.

\begin{equation}
    \label{eqn:attention}
    d' = e \cdot \sigma(S(e || d)) + d \cdot (1 - \sigma(S(e || d)))
\end{equation}

where $(e || d)$ refers to concatenation and $\sigma$ is the sigmoid function.
In the skip network architecture, we aim to compute an attention mask that reflects a maximal understanding of the image. In order to achieve this, we use an architecture inspired by U-Net\cite{ronneberger2015u}, which downscales the concatenated features to 256x8x8. By doing this, we achieve a mask computed with a considerable amount of receptive field while preserving the input features with the residual connections. We use the residual blocks from the original generator to upsample and downsample features in the skip network \cite{dalva2022vecgan} with channel size 256.
The skip network also takes both encoded and decoded features, and so it is equipped to extract the information of which attributes are edited when calculating which parts should be taken from the encoded features from the original image and which parts should be taken from the edited and decoded features.

\subsection{Training pathways}

We train our network using two different paths by modifying the translation paths defined by \cite{li2021image}. For each iteration to optimize our model, we sample a tag $i$ for shift direction, a source attribute $j$ as the current attribute, and a target attribute $\hat{j}$.

\noindent \textbf{Non-translation path.}
To ensure that the encoder-decoder structure preserves the images' details, we reconstruct the input image without applying any style shifts.
The resulting image is denoted as $x_n$ as given in Eq. \ref{eqn:non-tr}. 
\begin{equation}
    x_n = G(E(x))
    \label{eqn:non-tr}
\end{equation}

\noindent \textbf{Cycle-translation path.}
We apply a cyclic translation to ensure we get a reversible translation from a latent guided scale.
In this path,  we first apply a style shift by sampling $z \in \mathcal{U}[0,1)$ and obtaining target $\alpha_t$ with $M_{i,\hat{j}}(z)$ for target attribute $\hat{j}$.
The translation uses $\alpha$ that is obtained by subtracting $\alpha_t$ from the source style.
The decoder generates an image, $x_t$, as given in Eq. \ref{eqn:cycle-t1}  where $e$ is encoded features from input image $x$, $e=E(x)$.

\begin{align}
    x_t = G(T(e, M_{i,j}(z) - P(e, i), i))\label{eqn:cycle-t1}
\end{align}

Then by using the original image, $x$, as a reference image, we aim to reconstruct the original image by translating $x_t$.
Overall, this path attempts to reverse a latent-guided style shift with a reference-guided shift.
The second translation is given in Eq. \ref{eqn:cycle-tr} where $e_t=E(x_t)$.

\begin{align}
    x_c = G(T(e_t, P(e, i) - P(e_t, i), i))\label{eqn:cycle-tr}
\end{align}

In our learning objectives, we use $x_n$ and $x_c$ for reconstruction and $x_t$ and $x_c$ for adversarial losses, and $M_{i,j}(z)$ for the shift reconstruction loss. Details about the learning objectives are given in the next section.

\subsection{Learning objectives}
\label{sec:objectives}

Given an input image $x_{i,j} \in \mathcal{X}_{i,j}$, where $i$ is the tag to manipulate and $j$ is the current attribute of the image, we optimize our model with the following objectives. In our equations, $x_{i,j}$ is shown as $x$.

\textbf{Adversarial Objective.}
We learn a discriminator employing an architecture with decreasing resolution and increasing channel size. Like the generator, we build our discriminator with channel sizes of \{32, 64, 128, 256, 512, 512, 512, 1024, 2048\}, reducing the feature map dimensions to 1x1. We concatenate the extracted style $\alpha_t$ from the input image to this latent code and apply a 1x1 convolution. This final convolution is specific to each tag-attribute pair so that the model can use this information.

During training, our generator performs a style shift either in a latent-guided or a reference-guided way, resulting in a translated image. 
In our adversarial loss, we receive feedback from the two steps of the cycle-translation path.
As the first component of the adversarial loss, we feed a real image $x$ with tag $i$ and attribute $j$ to the discriminator as the real example.
To give adversarial feedback to the latent-guided path, we use the intermediate image generated in the cycle-translation path, $x_t$.
Finally, to provide adversarial feedback to the reference-guided path, we use the final outcome of the cycle-translation path $x_c$.
Only $x$ acts as a real image; both $x_t$ and $x_c$ are translated images and are treated as fake images with different attributes.
The discriminator aims to classify whether an image is real or fake, given its tag and attribute. 
The objective is given as:

\begin{equation}
\begin{split}
    \mathcal{L}_{adv} = 2log(D_{i,j}(x)) 
    +  log(1 - D_{i, \hat{j}} (x_t)) \\
    +  log(1 - D_{i, j} (x_c))
    \label{eqn:adv}
\end{split}
\end{equation}

\textbf{Shift Reconstruction Objective.}
As the cycle-consistency loss performs reference-guided generation followed by latent-guided generation, we utilize a loss function to make these two methods consistent with each other \cite{lee2018diverse,huang2018multimodal,li2019attribute,li2021image}.
Specifically, we would like to obtain the same target scale, $\alpha_t$, both from the mapping and the encoded reference image generated by the mapped $\alpha_t$. The loss function is given in Eq. \ref{eqn:l_shift}.
\begin{equation}
    \mathcal{L}_{shift} = ||M_{i,j}(z) - P(e_t, i)||_1
    \label{eqn:l_shift}
\end{equation}
Those parameters, $M_{i,j}(z)$ and $P(e_t, i)$, are calculated for the cycle-translation path as given in Eq. \ref{eqn:cycle-t1} and \ref{eqn:cycle-tr}.

\textbf{Image Reconstruction Objective.}
In all of our training paths, the purpose is to be able to re-generate the original image again.
To supervise this desired behavior, we use $L_1$ loss for reconstruction loss.
In our formulation, $x_n$ and $x_c$ are outputs of the non-translation and cycle-translation paths, respectively. Formulation of this objective is provided in Eq. \ref{eqn:rec_loss}.

\begin{equation}
    \begin{split}
        \mathcal{L}_{rec} = ||x_n - x||_1 + 
        ||x_c - x||_1
    \end{split}
    \label{eqn:rec_loss}
\end{equation}

\textbf{Orthogonality Objective.}
 To encourage the orthogonality between directions, we use soft orthogonality regularization based on the Frobenius norm, which is given in Eq. \ref{eqn:ortho_loss}.
 This orthogonality further encourages a disentanglement in the learned  style directions.
 
\begin{equation}
\mathcal{L}_{ortho} =  {\|A^{T}A - I\|_F} 
  \label{eqn:ortho_loss}
\end{equation}

\textbf{Disentanglement Objective}
We intend to change the scale for the desired semantic in each translation. As a reflection of this criteria, we penalize the changes in the attributes that are not subjected to any translation. For translated tag $i$, input scales which we refer to as source scale $\alpha_s$, and edited scales $\alpha_s'$, the disentanglement loss is formulated in equation \ref{eqn:disen_loss}. In the formulation, $\alpha^k$ represents the semantic scale for tag $k$.
Scales are calculated based on the projection of features given in Eq. \ref{eqn:proj}.

\begin{equation}
    \mathcal{L}_{dis} = \sum_{k \neq i} ||\alpha_s^k - \alpha_s^{k}{'}||
    \label{eqn:disen_loss}
\end{equation}

We find this disentanglement to be complementary to our orthogonality objective. When the model is trained with additional disentanglement loss, we observe that orthogonality loss drops to a lower value.

\textbf{Full Objective.}
Combining all of the loss components described, we reach the overall objective for optimization as given in Eq. \ref{eqn:full_loss}.
Additionally, we add an $L_1$ loss on the matrix $A$ parameters to encourage its sparsity.

\begin{equation}
    \begin{split}
        \underset{E,G,M,A}{\min} \underset{D}{\max} \lambda_{a}\mathcal{L}_{adv} +  \lambda_{s} \mathcal{L}_{shift} + \lambda_{r} \mathcal{L}_{rec} \\ +
        \lambda_{o} (\mathcal{L}_{ortho} + \mathcal{L}_{dis})+
        \lambda_{sp} \mathcal{L}_{sparse}
    \end{split}
    \label{eqn:full_loss}
\end{equation}

We set the following parameters; $\lambda_{a} = 1$, $\lambda_{rec} = 1.5$, $\lambda_{s} = 1$, $\lambda_{o} = 1$ and $\lambda_{sp} = 0.1$. 
We use a  learning rate of $10^{-4}$ and train our model for 600K iterations with a batch size of 8 on a single GPU with Adam optimizer.
Training takes 3 days.

\section{Experiments}
\label{sec:exp}

\subsection{Dataset and Settings}

We train our model on CelebA-HQ \cite{celeba} which contains 30,000 face images.
To extensively compare with state-of-the-arts, we use two train-evaluation protocols as follows:

\textbf{Setting I.} In our first setting, we follow the set-up from HiSD \cite{li2021image} to compare our method with end-to-end based image translation algorithms. Following HiSD, we use the first 3000 images of the CelebA-HQ dataset as the test set and 27000 as the training set.
These images include annotations for different attributes from which we use hair color, the presence of glass, and bangs attributes for translation tasks in this setting as in HiSD \cite{li2021image}.
The images are resized to $128\times128$.
Following the evaluation protocol proposed by HiSD \cite{li2021image}, we compute FID scores on the bangs addition task.
For each test image without bangs, we translate them to images with bangs with latent and reference guidance.
In latent guidance, 5 images are generated for each test image by randomly sampled scales from a uniform distribution.
This generated set of images is compared with images with attribute bangs in terms of their FIDs.
FIDs are calculated for these 5 sets and averaged. 
We randomly pick 5 reference images for reference guidance to extract the style scale.
FIDs are calculated for these 5 sets separately and averaged.

\textbf{Setting II.} We use our second setting to comprehensively compare our method with StyleGAN2-based inversion and editing methods. For this setting, the training/test split is obtained by re-indexing each image in CelebA-HQ back to the original CelebA and following the standard split of CelebA. This results in 27,176 training and 2,824 test images.
A single model is trained for hair color, the presence of glasses,  bangs, age, smiling, and gender attributes.
Images are resized to $256\times256$ resolution, which is the dimension StyleGAN2-based inversion methods use.
We evaluate our model with smile addition and removal and bangs addition attributes. Since the task of smile addition/removal requires a high-level understanding of the input face for modifying multiple facial components simultaneously, it is considered one of the most challenging attributes to edit. By benchmarking our model with such an editing task, we demonstrate the effectiveness of our framework in terms of image understanding.

\begin{table}[t]
    \centering
    \begin{tabular}{|l|c|c|}
    \hline
      \textbf{Method} &  \textbf{Lat.} & \textbf{Ref.}\\
\hline
VecGAN  \cite{dalva2022vecgan} w/o Orthogonality & 21.98 & 22.50 \\
VecGAN  \cite{dalva2022vecgan} w/o Sparsity & 24.07 & 22.43\\
VecGAN  \cite{dalva2022vecgan} &  20.17 & 20.72 \\
\hline
 Disent. & 20.23 & 20.57 \\
Disent. + Attn. Skip  & 20.15 & 20.08\\
Disent. +  Attn. UNet Skip &  19.98 & 19.87  \\
Disent. + Attn. UNet Skip w/ dec.  & \textbf{19.65} & \textbf{19.62}  \\
  \hline
    \end{tabular}
    \caption{Ablation Study on Setting
 I. Lat: Latent guided, Ref: Reference guided.}
    \label{tab:abl}
\end{table}

\begin{figure}
    \centering
    \includegraphics[width=1.0\textwidth]{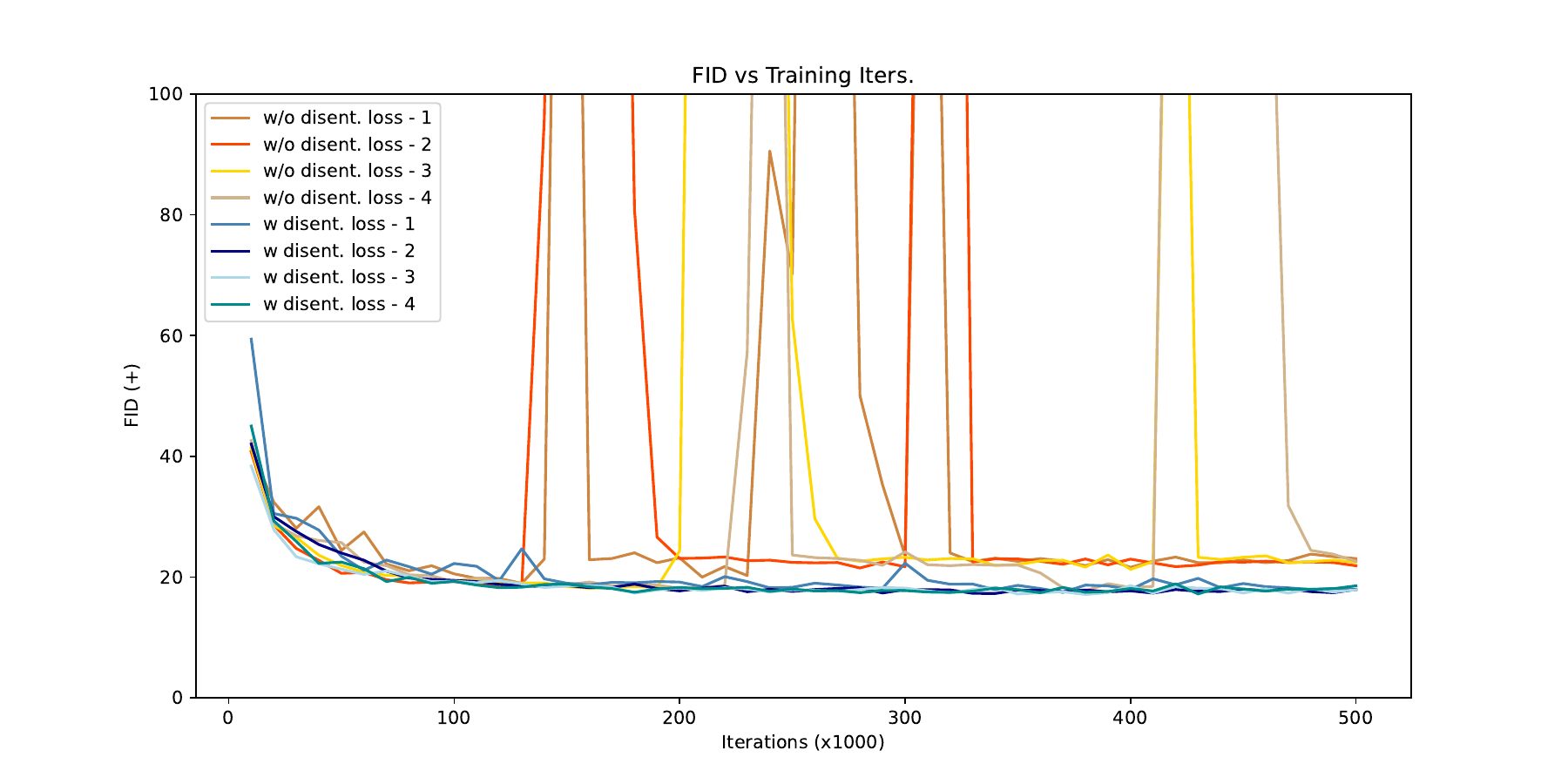}
    \caption{FID curves on smile addition of models trained with and without disentanglement loss through iterations.}
    \label{fig:fid-stability}
\end{figure}

\newcommand{\interpfigat}[1]{\includegraphics[trim=0 0 0cm 0, clip, width=2cm]{#1}}

\begin{figure}[t]
\centering
\scalebox{0.71}{
\addtolength{\tabcolsep}{-5pt}   
\begin{tabular}{ccccc}
\\
& \multicolumn{2}{c}{Hair Change} & \multicolumn{2}{c}{Smile Change}
\\
\interpfigat{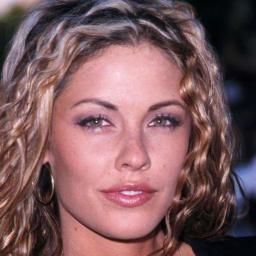} &
\interpfigat{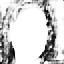} &
\interpfigat{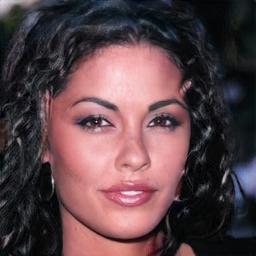} &
\interpfigat{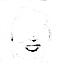} &
\interpfigat{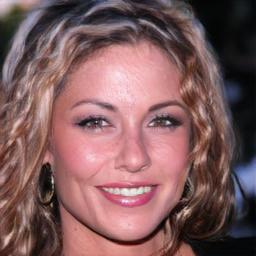} 
\\
\interpfigat{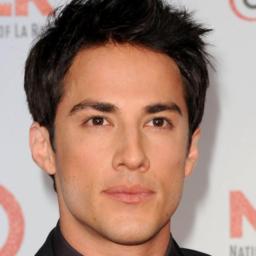} &
\interpfigat{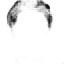} &
\interpfigat{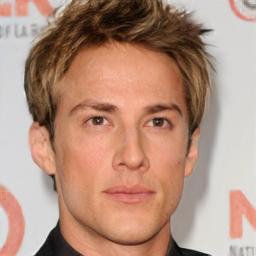} &
\interpfigat{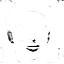} &
\interpfigat{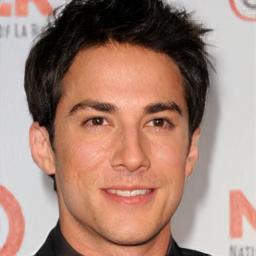} 
\\
Input & Attention Mask & Output & Attention Mask & Output\\
\end{tabular}
}
\caption{Visualizations of attention masks for different edits.
}
\label{fig:attn}
\end{figure}

\begin{figure*}[t]
    \centering
    \includegraphics[width=1.0\textwidth]{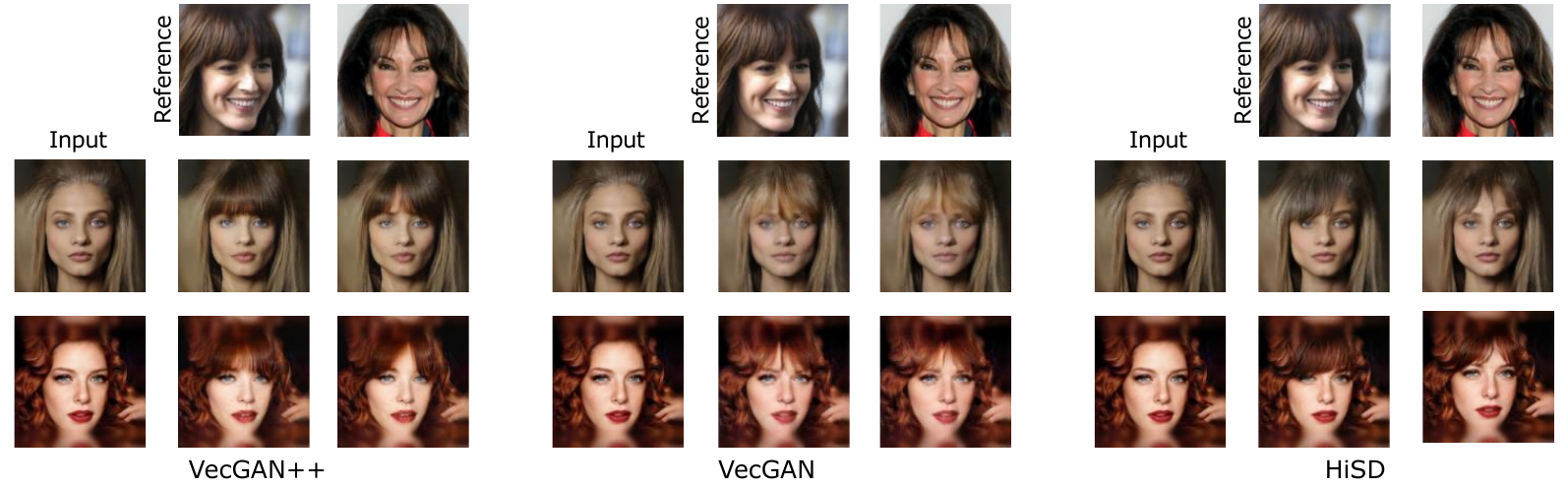}
    \caption{Qualitative results of bangs attribute of VecGAN++,  VecGAN and HiSD. 
    Given reference images, methods extract reference attributes and edit input images accordingly. 
    VecGAN++ achieves better edit quality compared to VecGAN and preserves  the other facial details better.
    It is important to note that, HiSD learns feature-based local translators, which is a successful approach on local edits, e.g. bangs, eyeglasses, hair color but not smile, age, or gender.
    Our method achieves comparable visual and better quantitative results than HiSD on this local task and can also achieve global edits.}
    \label{fig:hisd_comp}
\end{figure*}

\begin{table}[t]
    \centering
    \begin{tabular}{|l|c|c|c|}
    \hline
      \textbf{Method} &  \textbf{Latent} & \textbf{Reference} & \textbf{Avg.}\\
\hline
SDIT \cite{wang2019sdit} & 33.73 & 33.12 & 33.42\\
StarGANv2 \cite{starganv2} & 26.04 & 25.49 & 25.77 \\
Elegant \cite{xiao2018elegant} & - & 22.96 &  -  \\ 
HiSD \cite{li2021image} &  21.37 &  21.49 & 21.43 \\
VecGAN  \cite{dalva2022vecgan} &  20.17 & 20.72 & 20.45 \\
\hline
VecGAN++ & \textbf{19.65} & \textbf{19.62} & \textbf{19.64} \\
  \hline
    \end{tabular}
    \caption{Quantitative results for Setting
 I.}
    \label{table:results_setA}
\end{table}

\subsection{Metrics}
\label{sec:metrics}

We mostly build our evaluation on the FID metric as in previous works. Additionally, for smile manipulation, we evaluate our results on other metrics such as smile classification accuracy, Identity Preservation, and Background Preservation, described as follows:

\textbf{Frechet Inception Distance (FID)}: For the FID metric \cite{heusel2017gans}, we calculated the distance between the feature vectors of original and generated images, which are obtained using the Inception-V3 model. 
We set-up the FID evaluation based on the attributes that are edited. For example, for smile addition attribute edit, we edit the images that have negative smile tag from the validation set, those become our generated images. 
For the ground-truth distribution, we filter the images for the ones that have positive smile tag. 
Therefore, for the FID to be lower, the edit needs to be applied since our source images come from non-smiling images and target distribution images are smiling ones.

\textbf{Accuracy (Acc)}: We train an image classification network for attributes of faces on the training split of the CelebA-HQ dataset.
We use an ImageNet pretrained ResNet-50 model and fine-tune it for this task on 40 different attributes given by the dataset.
The model achieves $94\%$ accuracy on the validation set for the smile attribute.
We use this classifier to evaluate the accuracy of the generated images to test if the attribute is correctly manipulated. 

\textbf{Identity Preservation (Id)}: To calculate the Id metric, we use the CurricularFace model \cite{huang2020curricularface} to calculate the similarity between the original and generated images. The CurricularFace model uses the ResNet-101 model as a backbone for the feature extraction. 
We calculate the cosine similarity score  between the features of edited and original images.

\textbf{Background Preservation (BG)}: For the BG metric, we first use the facial attribute segmentation masks from the CelebAMask-HQ dataset to form background masks. Using these masks, we calculate the mean structural similarity index between the backgrounds of the original and edited images.

\subsection{Ablation Study} 
\label{sec:abl_results}

We provide an ablation study in Table \ref{tab:abl}. We report results without orthogonality and sparsity losses and then start from VecGAN and add the proposed disentanglement loss.
We observe improvements on FIDs as well as training stability with the disentanglement loss. We show this behavior in Fig. \ref{fig:fid-stability} by plotting training curves of models trained with and without disentanglement loss for 4 different runs which are initialized with different random seeds.

Our previous work, VecGAN has skip connection from encoder to the decoder to reconstruct the fine details from the input image.
In this work, we introduce a selection mechanism in the skip connections to ease the learning for the network.
We experiment with different variants of the attention based skip layers in Table \ref{tab:abl}.
We first experiment with attention based feature summation as given in in Eq. \ref{eqn:attention} but without the UNet architecture instead with a single layer module as well as without feeding the decoded features to the layer. 
As shown in the third row of the Table \ref{tab:abl}, results already start to improve.
Next, we increase the receptive field of the module by replacing it with a UNet architecture as explained in the Section \ref{sec:skip}.
Finally, we also feed the decoded features to the UNet so that the model can explore what is edited and what the input is to select which features should be decoded for high quality edits.
These quantitative results are averaged between 5 runs.
We also visualize the attention masks of different edits in Fig. \ref{fig:attn}.
Attention masks detect edited areas as the module receives information from both encoded and edited features.
Thanks to these attention masks, unedited pixels are passed to the decoder for high-fidelity reconstruction.

\subsection{Comparisons with Competing Methods} 
\label{sec:results}

We extensively compare our results with other end-to-end image translation methods.
In Setting I, as given in Table \ref{table:results_setA}, we compare with SDIT \cite{wang2019sdit}, StarGANv2 \cite{starganv2}, Elegant \cite{xiao2018elegant}, and HiSD \cite{li2021image} models.
Among these methods, HiSD learns a hierarchical style disentanglement, whereas StarGANv2 learns a mixed style code.
Therefore StarGANv2, when translating images, also edits other attributes  and does not strictly preserve the identity.
HiSD achieves disentangled style edits.
However, HiSD learns feature-based local translators, an approach known to be successful on local edits, e.g. bangs, and their model is trained for bangs, eyeglasses, and hair color attributes.
VecGAN achieves significantly better quantitative results than HiSD both in latent-guided and reference-guided evaluations, even though they are compared on a local edit task.
We further achieve improvements with VecGAN++.

Fig. \ref{fig:hisd_comp} shows reference-guided results of our final model, VecGAN++, VecGAN, and HiSD.
As shown in Fig. \ref{fig:hisd_comp},  methods achieve attribute disentanglement, they do not change any other attribute of the image than the bangs tag.
VecGAN++ achieves better edit quality compared to VecGAN.
It is important to note that, HiSD learns feature-based local translators, which is a successful approach on local edits, e.g. bangs, eyeglasses, and hair color but not smile, age, or gender.
Our method achieves comparable visual and better quantitative results than HiSD on this local task and can also achieve global edits.

\begin{table*}[t]
\centering
\begin{tabular}{|l|c|c|c|c|c|c|c|c|c|c|c|}
\hline
& \multicolumn{2}{c|}{Smile} & \multicolumn{2}{c|}{Bangs} & \multicolumn{2}{c|}{Age} & \multicolumn{2}{c|}{Gender} & Blonde & Black & Brown \\
\hline
\textbf{Method} & (+) & (-) & (+) & (-) & (+) & (-) & Female & Male & (+) &  (+) & (+)  \\
\hline

e4e \cite{tov2021designing} & 35.01 & 37.91 & 53.29 & 53.62 & 52.28 & 60.38 & 36.55 & 62.27 & 69.26 & 64.92 & 40.93 \\

HyperStyle \cite{alaluf2022hyperstyle} & 25.25 & 24.64 & 41.37 & 47.32 & 52.23 & 44.60 & 42.77 & 62.11 & 60.50 & 59.63 & 35.31 \\
HFGI \cite{wang2022high} & 23.49 & 26.58 & 40.54 & 45.06 & 45.30 & 46.20 & 43.98 & 64.84 & 67.92 & 60.09 & 35.15 \\
StyleTran. \cite{Hu_2022_CVPR} & 27.64 & 32.71 & 44.66 & 51.91 & 55.24 & 55.67 & 46.34 & 57.81 & 55.96 & 66.64 & 35.15 \\
StyleRes \cite{pehlivan2022styleres} & 20.53 & 21.63 & 40.13 & 38.05 & 45.00 & 44.24 & 34.73 & 61.39 & 60.12 & 54.78 & 38.31 \\
VecGAN \cite{dalva2022vecgan}  & 17.70 & 20.26 & 36.47 & 35.16 & 40.65 & 37.28 & 63.48 & 53.45 & 68.89 & 68.74 & 35.50 \\
\hline
VecGAN++ & \textbf{17.43} & \textbf{18.24} & \textbf{27.09} & \textbf{31.63} & \textbf{35.14} & \textbf{25.22} & \textbf{30.24} & \textbf{52.56} & \textbf{37.18} & \textbf{49.48} & \textbf{34.98} \\
\hline
\end{tabular}
\caption{Quantitative results for Setting II.}
\label{table:results_setB}
\end{table*}

\newcommand{\interpfigt}[1]{\includegraphics[trim=0.1cm 0 1.5cm 0, clip, width=5.6cm]{#1}}
\begin{figure*}[t]
  \centering
  \begin{tabular}{ccc}
    \interpfigt{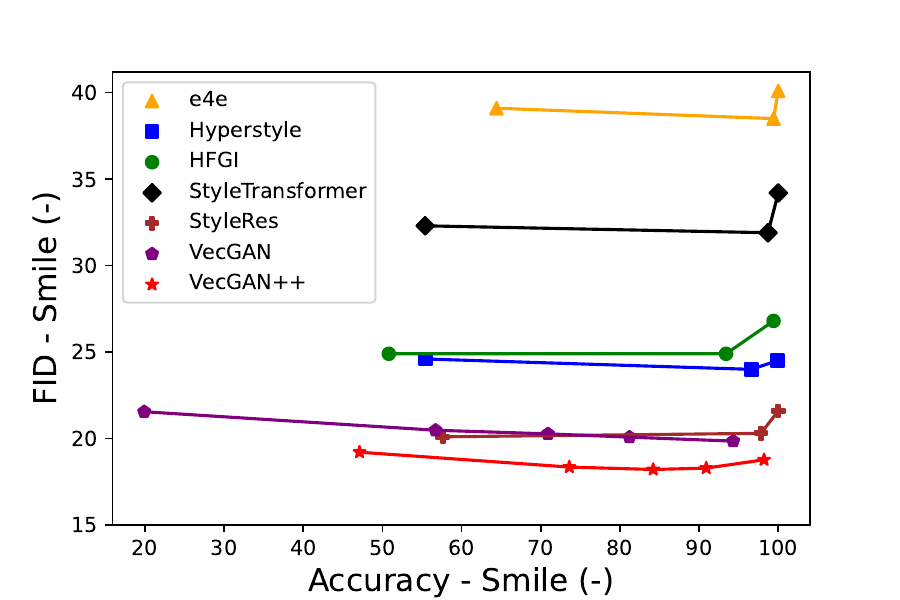}&
    \interpfigt{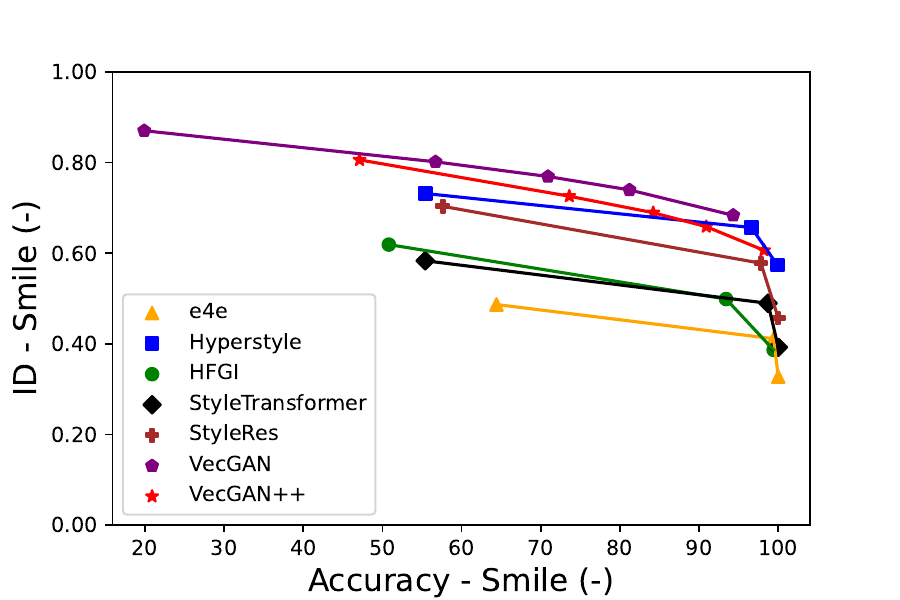}&
    \interpfigt{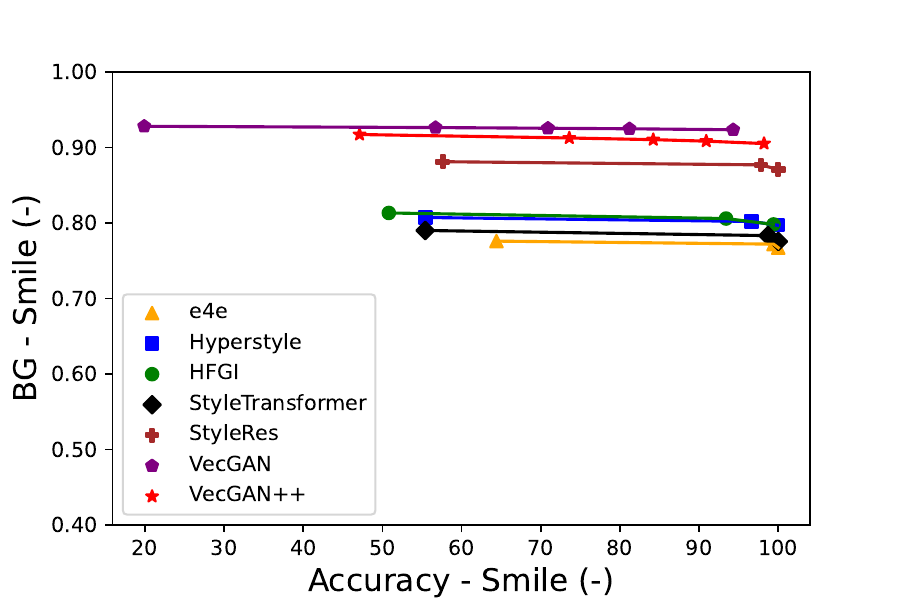} \\
    \interpfigt{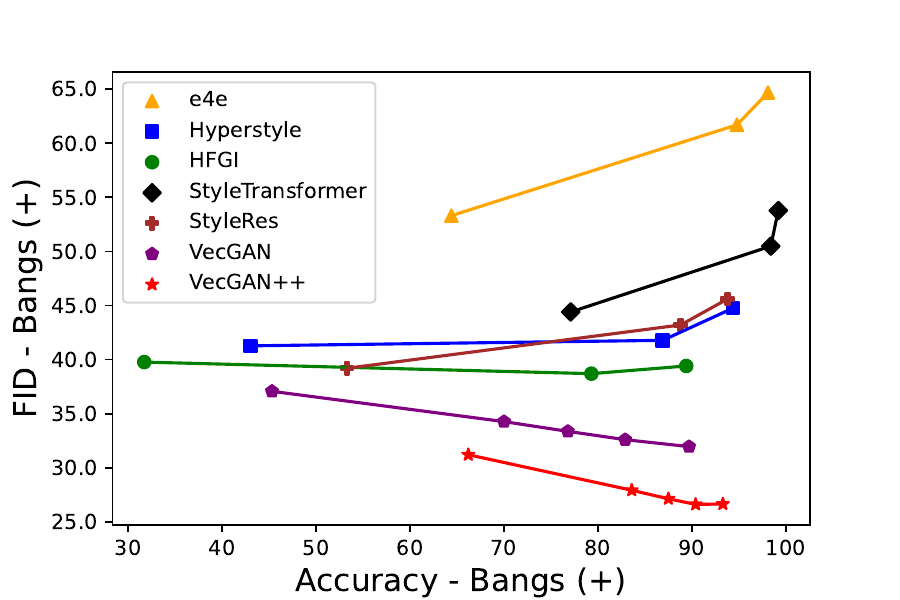}&
    \interpfigt{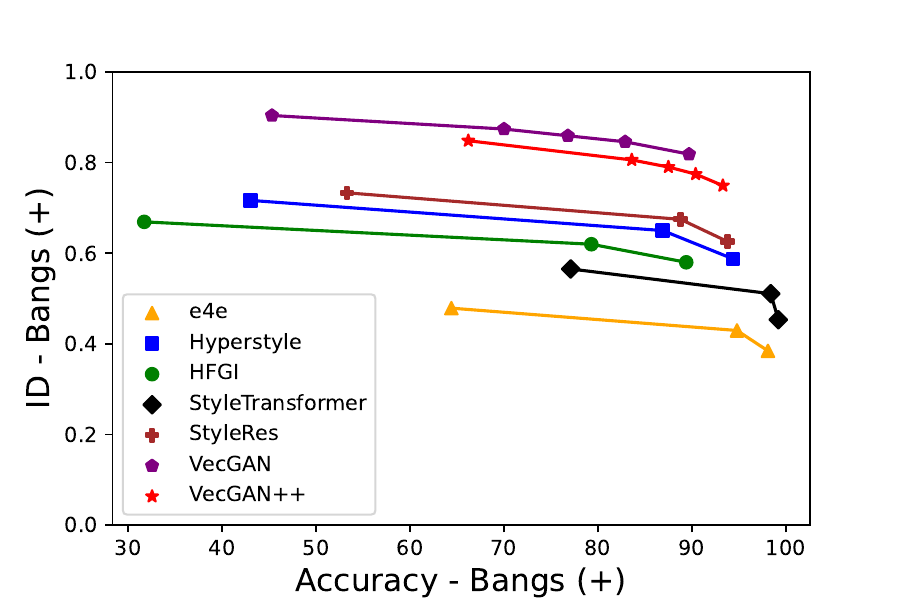}&
    \interpfigt{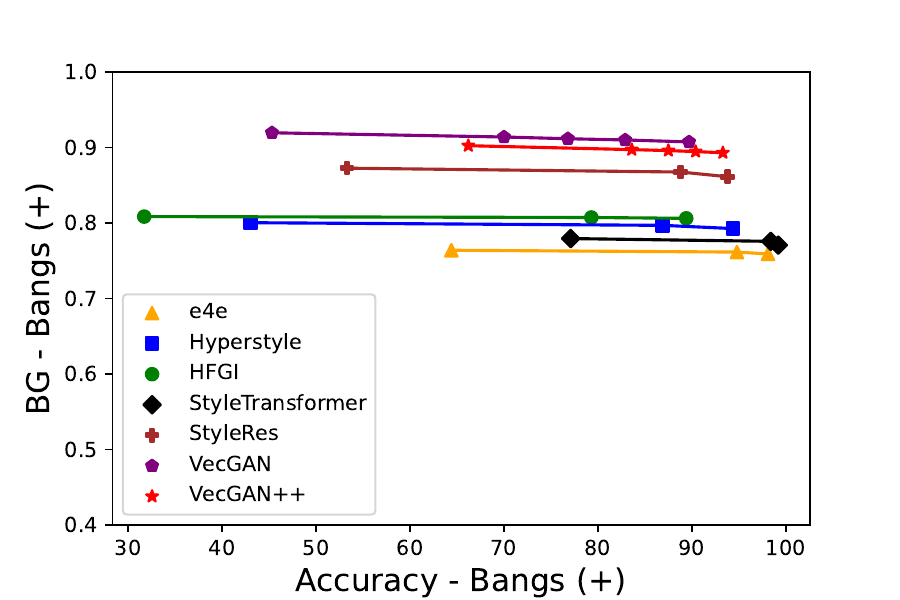} \\

     \end{tabular}
     \caption{Plots of FID, ID, and BG metrics as we change the intensity of the attributes and the number of steps to take for explored directions.  For each intensity, we measure the attribute accuracy in the x-axis. The first row plots present results for smile removal (global attribute), and the second row presents them for bangs addition (local attribute).}
     \label{fig:strength}
\end{figure*}

\newcommand{\interpfigk}[1]{\includegraphics[trim=0 0 0cm 0, clip, width=3cm]{#1}}

\begin{figure*}
\centering
\scalebox{0.71}{
\addtolength{\tabcolsep}{-5pt}   
\begin{tabular}{ccccccccc}
\\
\rotatebox{90}{~~~~~~1. Smile (+)} &
\interpfigk{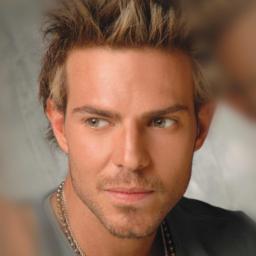} &
\interpfigk{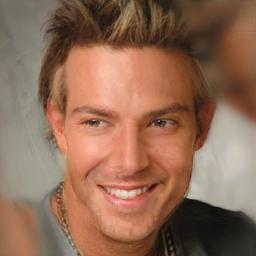} &
\interpfigk{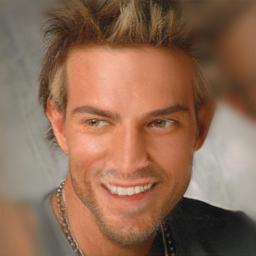} &
\interpfigk{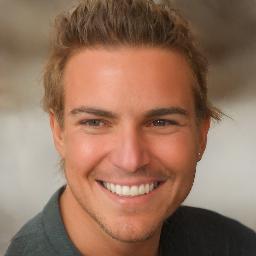} &
\interpfigk{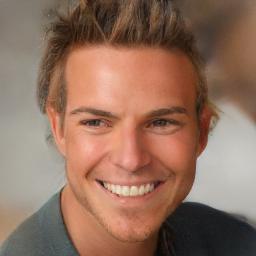} &
\interpfigk{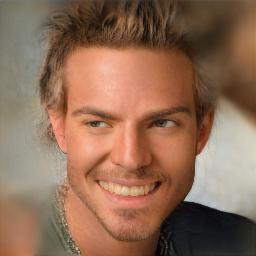} &
\interpfigk{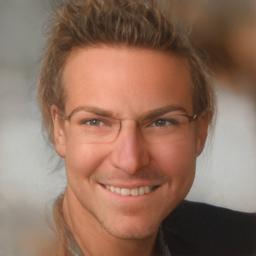} &
\interpfigk{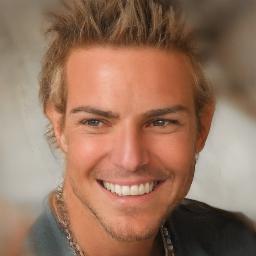}
\\
\rotatebox{90}{~~~~~~2. Smile (+)} &
\interpfigk{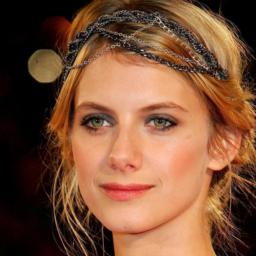} &
\interpfigk{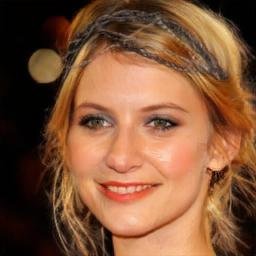} &
\interpfigk{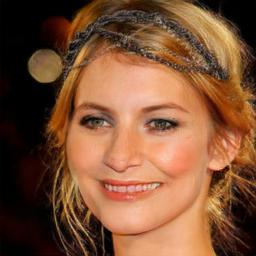} &
\interpfigk{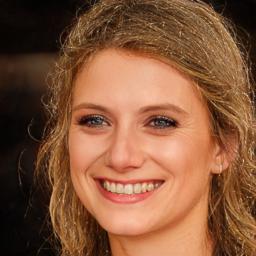} &
\interpfigk{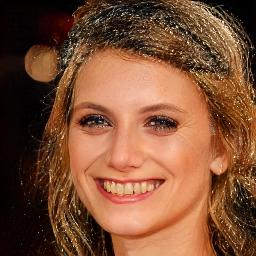} &
\interpfigk{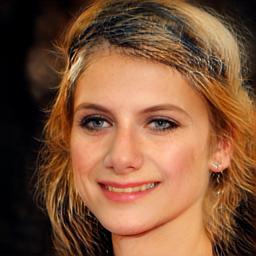} &
\interpfigk{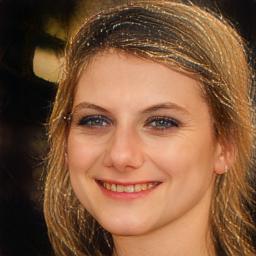}&
\interpfigk{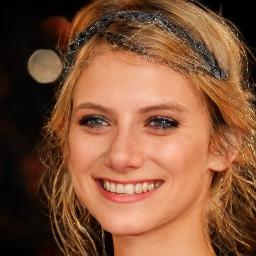}
\\
\rotatebox{90}{~~~~~~~3. Bangs (+)} &
\interpfigk{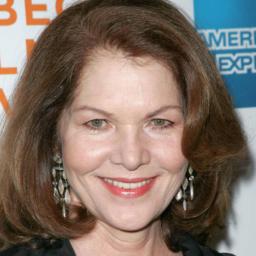} &
\interpfigk{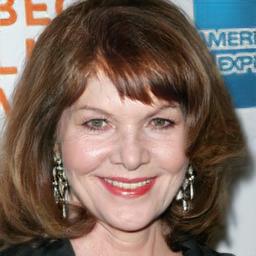} &
\interpfigk{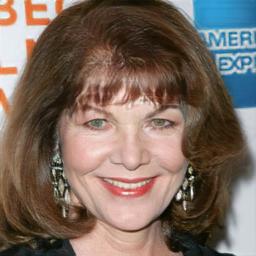} &
\interpfigk{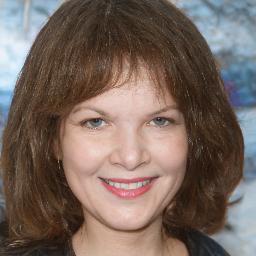} &
\interpfigk{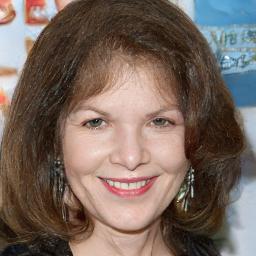} &
\interpfigk{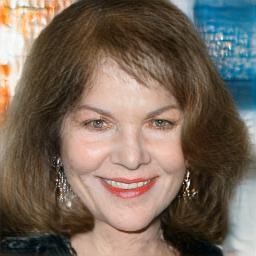} &
\interpfigk{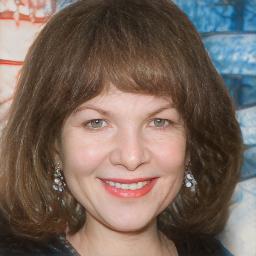} &
\interpfigk{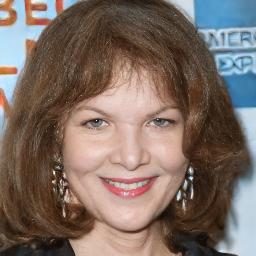}
\\
\rotatebox{90}{~~~~~~~4. Age (+)} &
\interpfigk{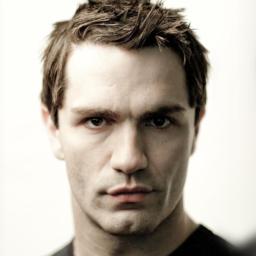} &
\interpfigk{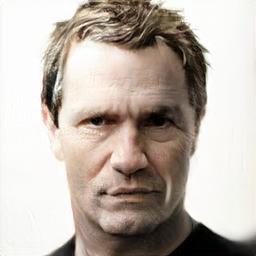} &
\interpfigk{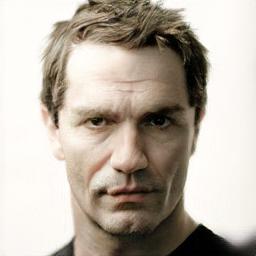} &
\interpfigk{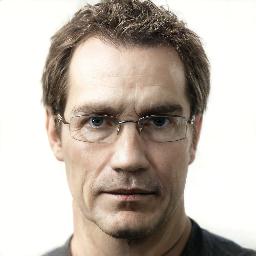} &
\interpfigk{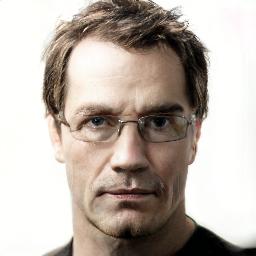} &
\interpfigk{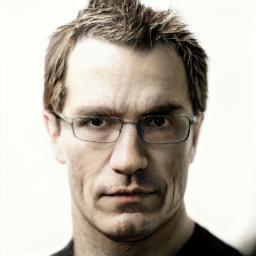} &
\interpfigk{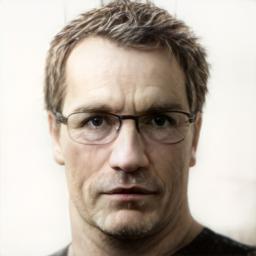} &
\interpfigk{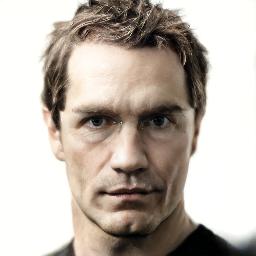}
\\
\rotatebox{90}{~~~~~~~~5. Male (+)} &
\interpfigk{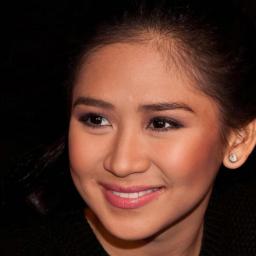} &
\interpfigk{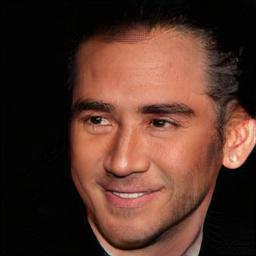} &
\interpfigk{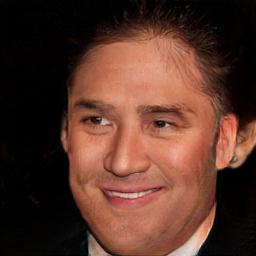} &
\interpfigk{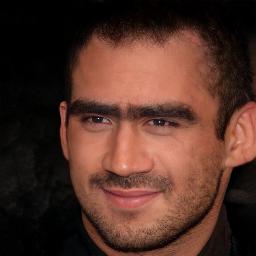} &
\interpfigk{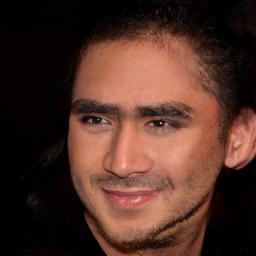} &
\interpfigk{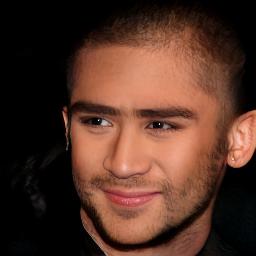} &
\interpfigk{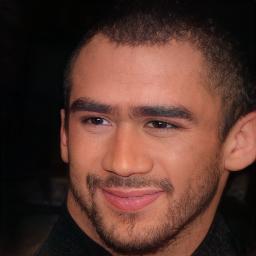} &
\interpfigk{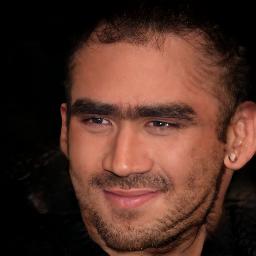}
\\
\rotatebox{90}{~~~~~~~~6. Female (+)} &
\interpfigk{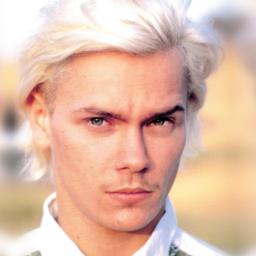} &
\interpfigk{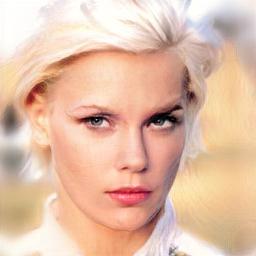} &
\interpfigk{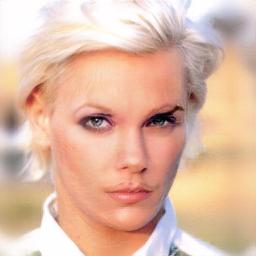} &
\interpfigk{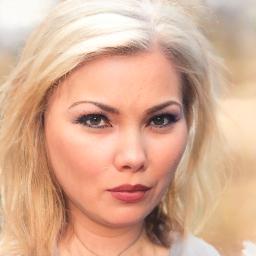} &
\interpfigk{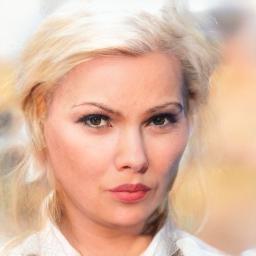} &
\interpfigk{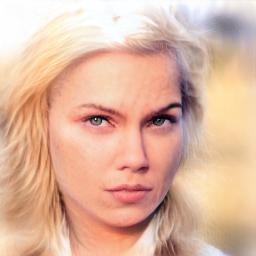} &
\interpfigk{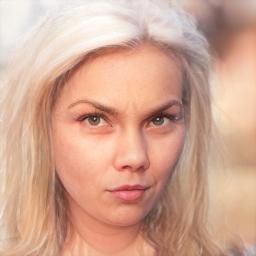} &
\interpfigk{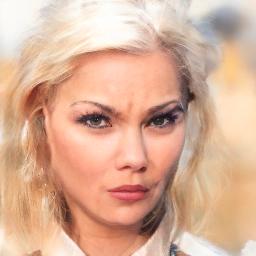}
\\
\rotatebox{90}{~~~~~~~~7. Blonde (+)} &
\interpfigk{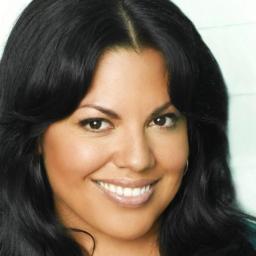} &
\interpfigk{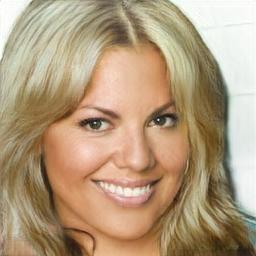} &
\interpfigk{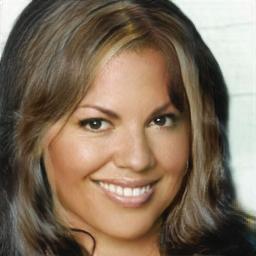} &
\interpfigk{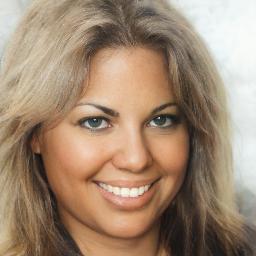} &
\interpfigk{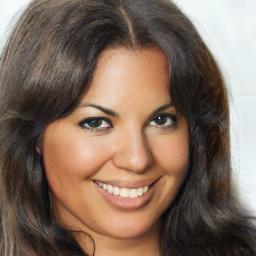} &
\interpfigk{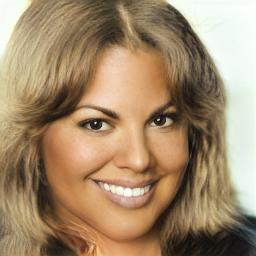} &
\interpfigk{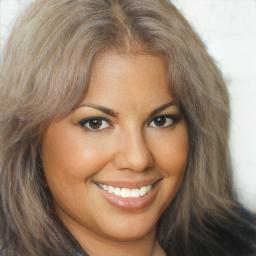} &
\interpfigk{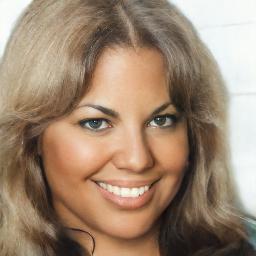}
\\
& Input & VecGAN++ & VecGAN & e4e & HFGI & HyperStyle & StyleTransformer & StyleRes \\
\end{tabular}
}
\caption{Qualitative results of our and competing methods. StyleGAN inversion-based methods do not faithfully reconstruct input images. 
StyleRes achieves better reconstruction results compared to others but not compared to VecGAN models.
VecGAN++ and VecGAN achieve  high fidelity to the originals with only targeted attributes manipulated naturally and realistically.
We also observe that VecGAN++ improves over VecGAN in many examples.
}
\label{fig:results_all}
\end{figure*}

In our second set-up of evaluation,
 we compare our method with state-of-the-art StyleGAN2 inversion-based methods, e4e \cite{tov2021designing}, HyperStyle \cite{alaluf2022hyperstyle}, HFGI \cite{wang2022high}, StyleTransformer \cite{Hu_2022_CVPR}, and StyleRes \cite{pehlivan2022styleres}  in Table \ref{table:results_setB}.
 We compare the methods on local 
 (e.g. bangs) and global attribute (e.g. smile) manipulations. 
 For the smile attribute, we use the direction explored by the InterfaceGAN method \cite{shen2020interpreting}.
 For the others, we use the direction discovered by the StyleCLIP method \cite{patashnik2021styleclip}.
 The strength attribute is set to the one that achieves the best FID scores. We provide more analysis on the strengths in Fig. \ref{fig:strength}.
We achieve significant improvements on the presented attributes both with respect to VecGAN and state-of-the-art StyleGAN2 inversion-based methods.

We also conduct a user study on 30 samples among 20 users. To generate 30 samples, first 6 images of validation set are used without cherry-picking and all 5 attribute edits are applied to these images.  We set an A/B test and provide users with input images and translated ones obtained by VecGAN++ and other competing methods, namely VecGAN and StyleRes.
We limit our user study to these two comparisons given the expenses of user study and since VecGAN and StyleRes significantly outperform the previous works as shown in Table \ref{table:results_setB} and  Fig. \ref{fig:results_all}. The left-right order is randomized to ensure fair comparisons.
We ask users to select the best result according to i) whether the selected attribute is correctly added, ii) 
whether irrelevant facial attributes preserved, iii) and overall whether the output image looks realistic and high quality.
We ask users to pay attention if details from the input image is preserved in addition to the quality and they are given the information of which attribute is targeted for edit for each sample.
Users select VecGAN++ as opposed to VecGAN $73.3\%$ of the time ($50\%$ is tie), and as opposed to StyleRes $67.3\%$ of the time.
These results are consistent with reported FID metrics and qualitative results.

Our method and  StyleGAN inversion-based methods provide a knob to control the editing attribute intensity.
We obtain plots provided in Fig. \ref{fig:strength} by changing the editing attribute intensity.
As we increase the intensity, edits become more detectable. We measure that with a smile classifier. Therefore, we plot FID, Id (Identity), and BG (Background reconstruction) scores with respect to the attribute intensity measured by the accuracy of the classifier.
Specifically, for VecGAN++ and VecGAN, we set $\alpha_t$ given in Eq. \ref{eqn:alphat} to $\{0.0, 0.33, 0.5, 0.66, 1.0\}$.
For StyleGAN inversion-based methods, we set the strength parameter to $\{1, 2, 3\}$.
These models usually set the strength to $3$ for successful smile edits.
As shown in Fig. \ref{fig:strength}, VecGAN++ and VecGAN achieve better FID scores compared to others consistently. 
With the highest strength, where the accuracy of the classifier goes to $100\%$ for all models, we observe that FID scores for StyleGAN inversion-based model scores drastically get worse, whereas our results are robust. VecGAN++ achieves better FID scores than VecGAN consistently. 
We find the Id score getting worse as the edit strength increases. 
That results from changes in the person and the limitations of the CurricularFace model.
VecGAN++ achieves significantly better BG scores than StyleGAN inversion-based models and slightly worse than VecGAN.
We observe that VecGAN++ achieves better edit quality, as reflected in FID scores, than VecGAN. On the other hand, VecGAN does less edits as measured by the accuracies and achieves slightly better Id and BG scores.

We provide qualitative comparisons in Fig. \ref{fig:results_all}.
StyleGAN inversion-based methods do not faithfully reconstruct input images. 
They miss many details from the background and foreground.
StyleRes achieves better reconstruction results compared to others but still worse than ours as measured by BG metrics. 
VecGAN++ and VecGAN achieve high fidelity to the originals with only targeted attributes manipulated naturally and realistically.
We also observe that VecGAN++ improves over VecGAN on many examples, especially on smile examples with less artefacts, a better quality bangs addition, and significant improvements on the gender manipulation.
Additionally, VecGAN++ achieves successful age edits, whereas StyleGAN inversion-based methods add eyeglasses very frequently.
This shows VecGAN++ and VecGAN provide with better disentanglement between correlated attributes, e.g. age and eyeglasses.
This is because our models are trained end-to-end with labeled datasets for this task. We provide more analysis on these comparisons in the next section.

\begin{figure*}
\begin{subfigure}[b]{0.32\linewidth}
    \centering
    \includegraphics[width=0.95\textwidth, trim={0 1.1cm 0 0},clip]{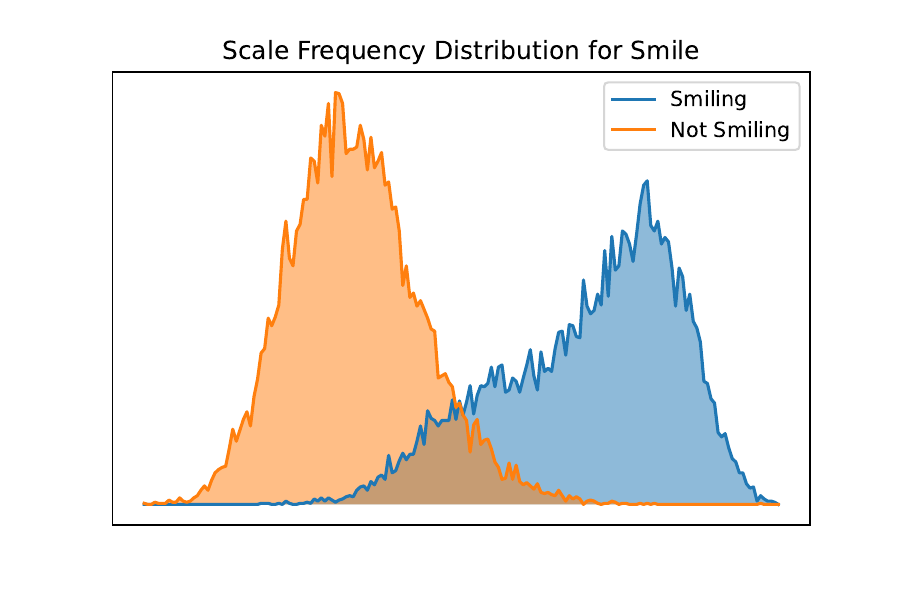}
    \captionsetup{width=0.95\textwidth}
      \caption{Histogram $\alpha_t$ values for smile tag of VecGAN}
    \label{fig:histog_smileV}
\end{subfigure}
   \begin{subfigure}[b]{0.32\linewidth}
    \centering
    \includegraphics[width=0.95\textwidth, trim={0 1.1cm 0 0},clip]{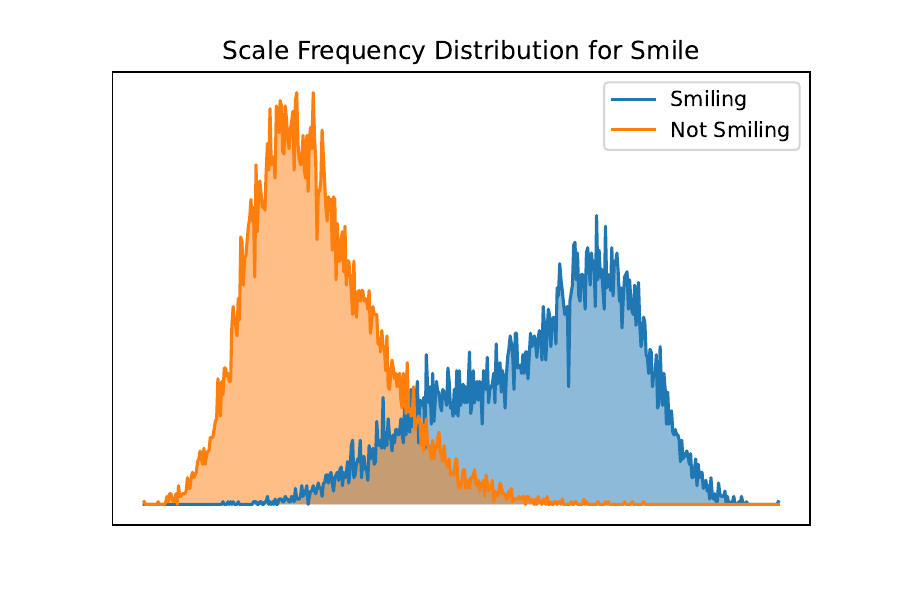}
    \captionsetup{width=0.95\textwidth}
      \caption{Historgram $\alpha_t$ values for smile tag of VecGAN++}
    \label{fig:histog_smile}
\end{subfigure}
    \begin{subfigure}[b]{0.32\linewidth}
    \centering
    \includegraphics[width=0.95\textwidth]{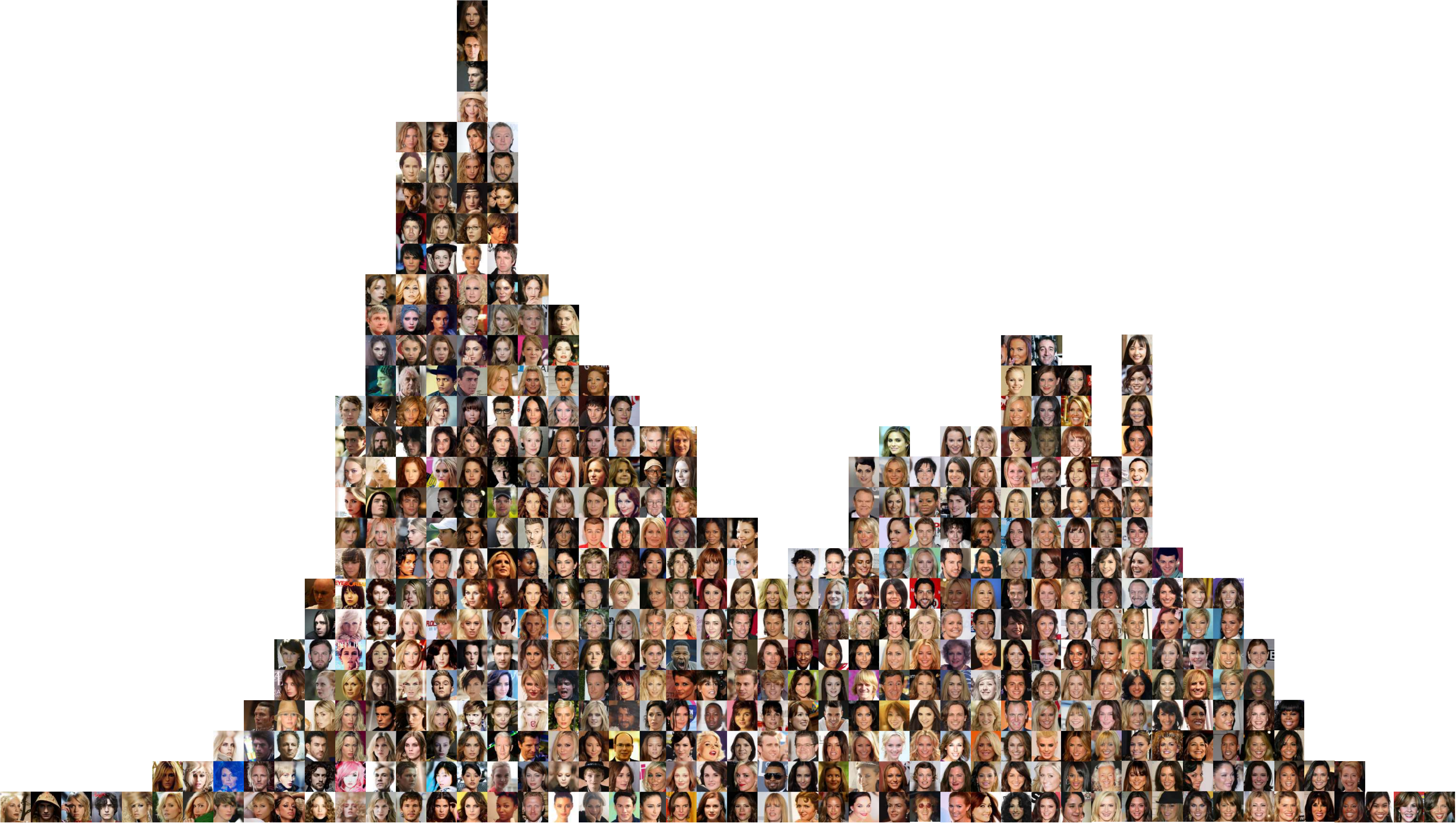}
    \captionsetup{width=0.95\textwidth}
    \caption{Training images plotted based on their $\alpha_t$ values from VecGAN++ for smile tag}
      \label{fig:smile_histo}
\end{subfigure}  
   \begin{subfigure}[b]{0.32\linewidth}
    \centering
    \includegraphics[width=0.95\textwidth, trim={0 1.1cm 0 0},clip]{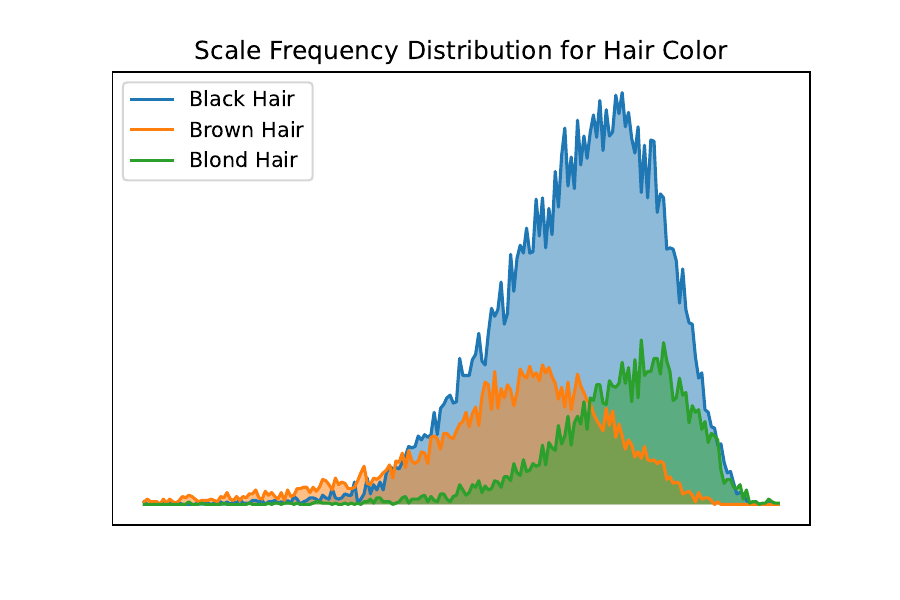}
    \captionsetup{width=0.95\textwidth}
      \caption{Histogram $\alpha_t$ values for hair color tag of VecGAN}
    \label{fig:histog_hairV}
\end{subfigure}
   \begin{subfigure}[b]{0.32\linewidth}
    \centering
    \includegraphics[width=0.95\textwidth, trim={0 1.1cm 0 0},clip]{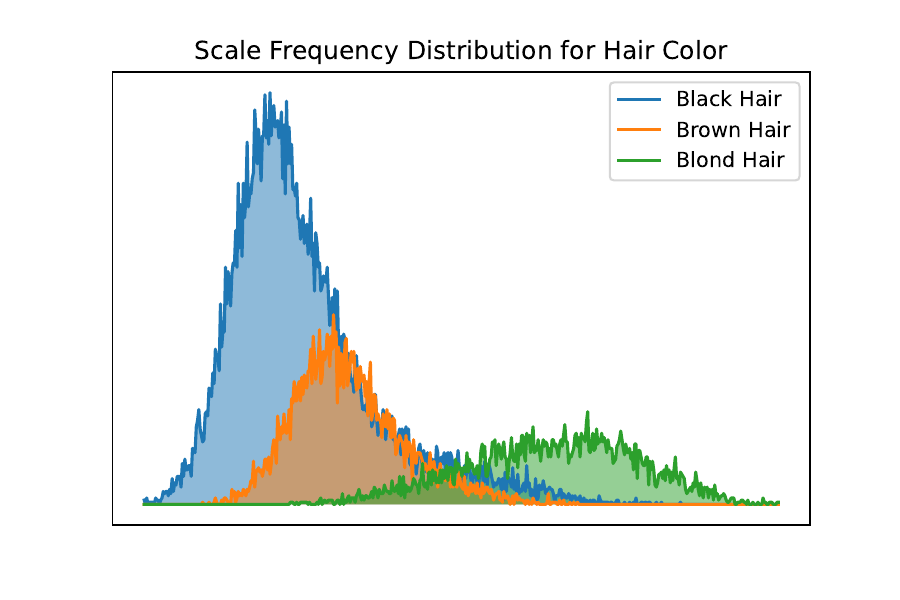}
    \captionsetup{width=0.95\textwidth}
      \caption{Histogram $\alpha_t$ values for hair color tag of VecGAN++}
    \label{fig:histog_hair}
\end{subfigure}
    \begin{subfigure}[b]{0.32\linewidth}
    \centering
    \includegraphics[width=0.95\textwidth]{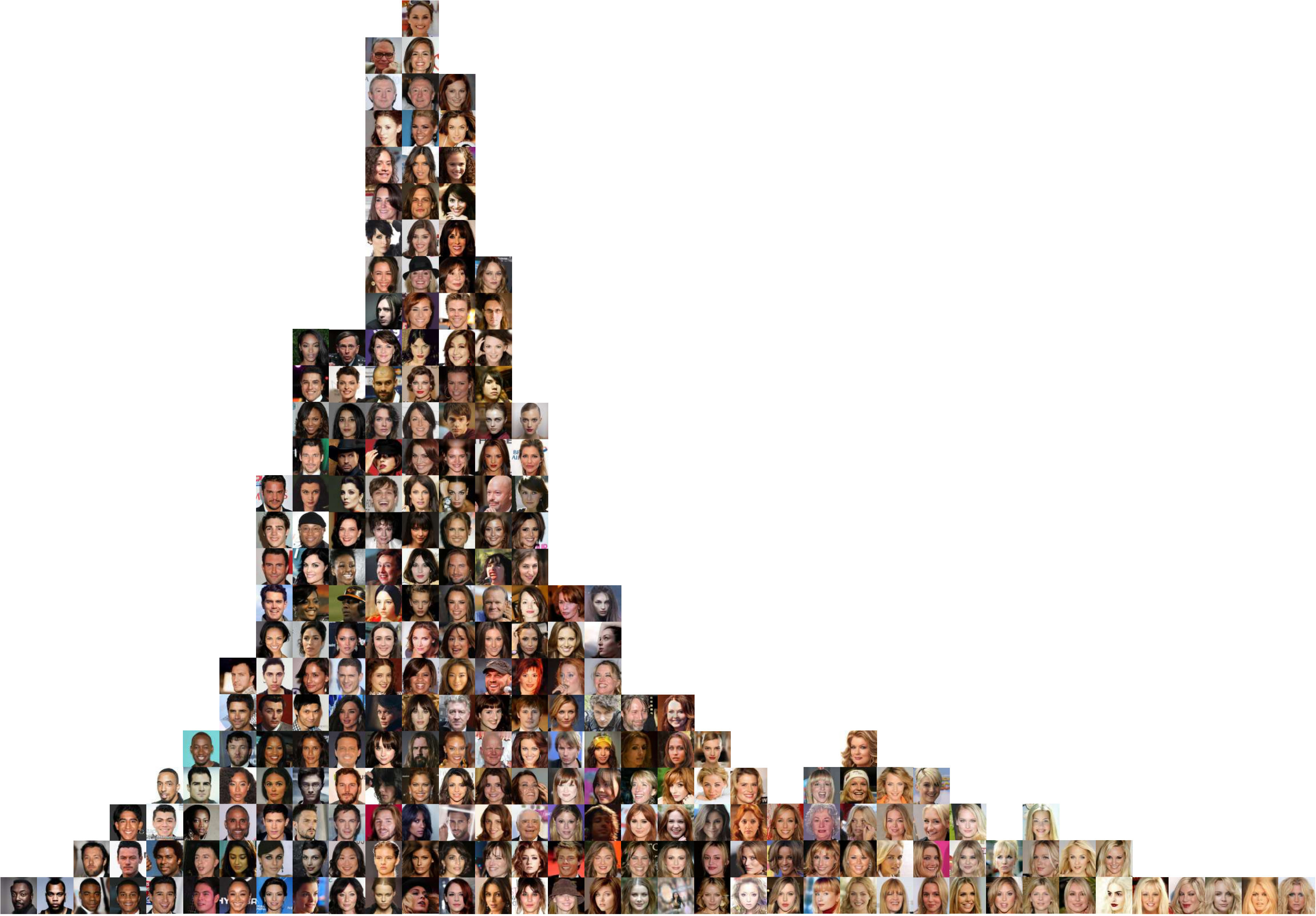}
    \captionsetup{width=0.95\textwidth}
    \caption{Training images plotted based on their $\alpha_t$ values from VecGAN++ 
 for hair color tag}
      \label{fig:hair_histo}
\end{subfigure}  
   \begin{subfigure}[b]{0.32\linewidth}
    \centering
    \includegraphics[width=0.95\textwidth, trim={0 1.1cm 0 0},clip]{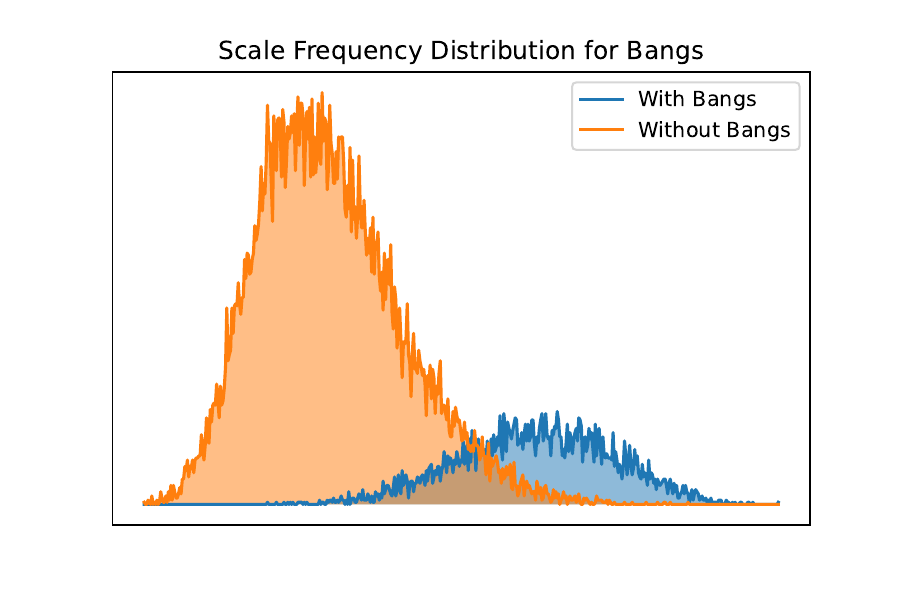}
    \captionsetup{width=0.95\textwidth}
      \caption{Histogram $\alpha_t$ values for bangs tag of VecGAN++}
    \label{fig:histog_bangs}
\end{subfigure}
   \begin{subfigure}[b]{0.32\linewidth}
    \centering
    \includegraphics[width=0.95\textwidth, trim={0 1.1cm 0 0},clip]{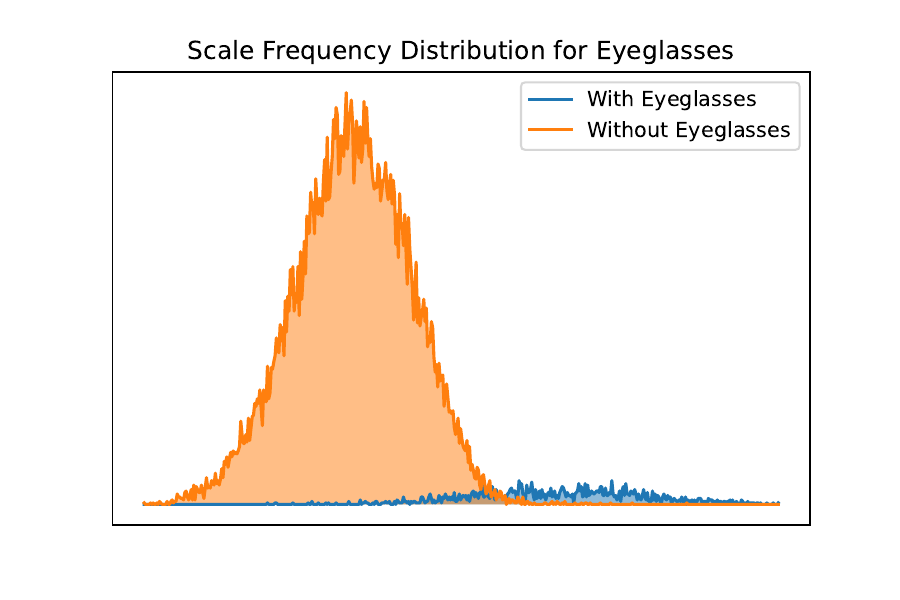}
    \captionsetup{width=0.95\textwidth}
      \caption{Histogram $\alpha_t$ values for eyeglasses tag of VecGAN++}
    \label{fig:histog_eyeglasses}
\end{subfigure}
    \begin{subfigure}[b]{0.32\linewidth}
    \centering
   \includegraphics[width=0.95\textwidth, trim={0 1.1cm 0 0},clip]{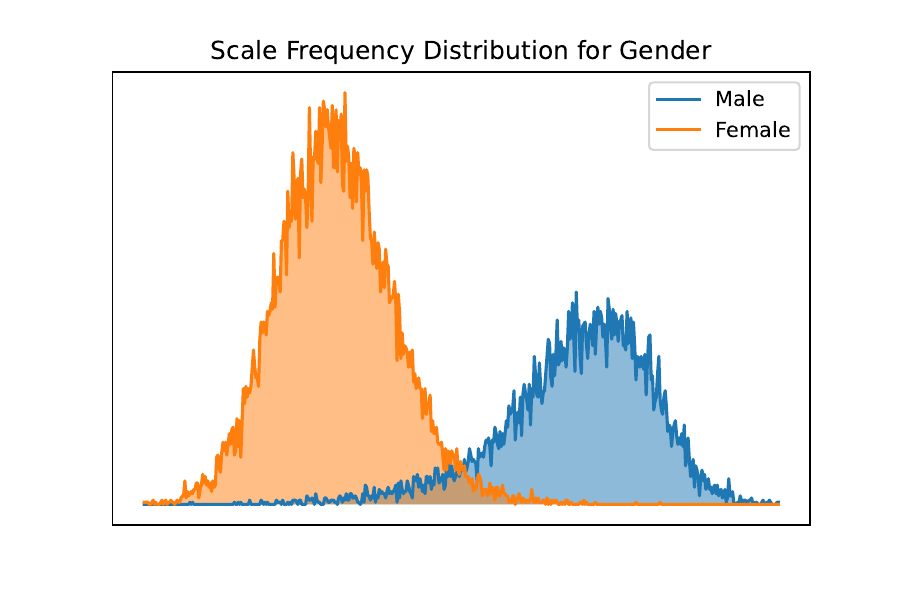}
    \captionsetup{width=0.95\textwidth}
    \caption{Histogram $\alpha_t$ values for gender tag of VecGAN++}
      \label{fig:gender_histo}
\end{subfigure}  
    \caption{Analysis of $\alpha_t$ for hair color and smile tags. For details in image histograms, please zoom in.}
    \label{fig:hair_histo_all}
\end{figure*}

\section{Analysis and Discussions}
\label{sec:analysis}

\subsection{Comparing End-to-end Image Translation Networks versus StyleGAN Inversion-based Methods}

We propose an end-to-end trained image translation network in this work and extensively compare our method with StyleGAN inversion-based methods.
We note the different advantages and disadvantages of both approaches.

We observe that end-to-end trained image translation networks, especially our proposed framework, do not suffer from the reconstruction and editability trade-off. This trade-off is pointed out for StyleGAN inversion-based methods \cite{tov2021designing}.
That is, when the inversion parameters are optimized to reconstruct the input images faithfully, they do not lie in the natural StyleGAN distribution space, and therefore the edit quality gets poor for those high-fidelity inversions. 
That is the advantage of our method because it is trained end-to-end, and we learn both reconstruction and editing together.

StyleGAN inversion-based methods enjoy many editing capabilities, whereas our framework only achieves pre-defined edits for which it is trained. Those methods that employ pre-trained StyleGANs rely on StyleGAN's semantically rich feature organizations. 
The editing directions are discovered after StyleGAN is trained.
Some methods discover directions in supervised and unsupervised ways.
Supervised methods, e.g. InterfaceGAN, require labeled datasets the same as ours. 
On the other hand, with unsupervised methods and text-based editing methods, directions are explored for those that do not have labeled datasets. 
For example, with the GANSpace method \cite{harkonen2020ganspace}, editing directions are found for different expressions, and with the StyleCLIP method \cite{patashnik2021styleclip}, editing directions are found for different hairstyles (Mohawk hairstyle, Bob-cut hairstyle, Afro hairstyle, e.g.).
That is an advantage of StyleGAN inversion-based methods.

\begin{figure*}
\centering
\scalebox{0.71}{
\addtolength{\tabcolsep}{-5pt}   
\begin{tabular}{ccccccccc}
\\
\interpfigk{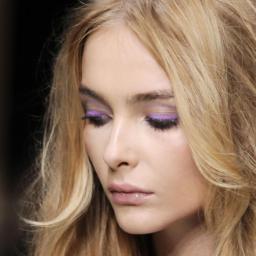} &
\interpfigk{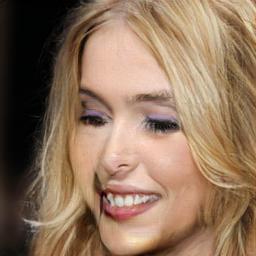} &
\interpfigk{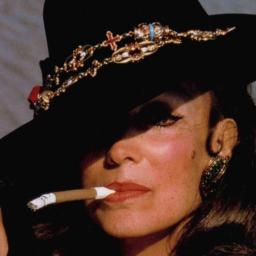} &
\interpfigk{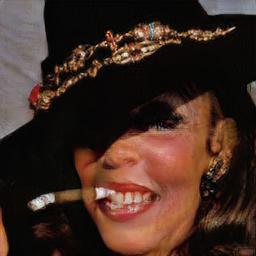} &
\interpfigk{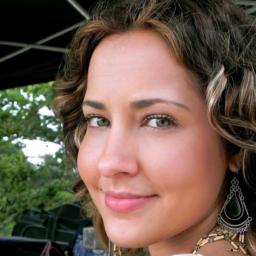} &
\interpfigk{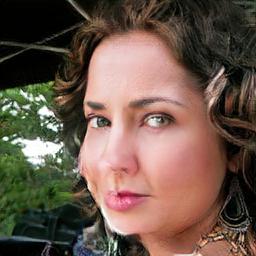} &
\interpfigk{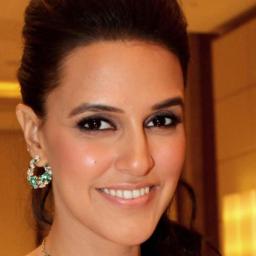} &
\interpfigk{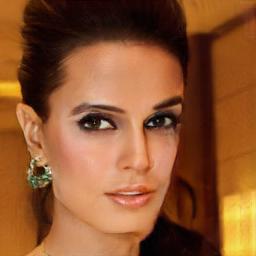} &
\\
Input & Smile (+) & Input & Smile (+) & Input & Smile (-) & Input & Smile (-) \\
\interpfigk{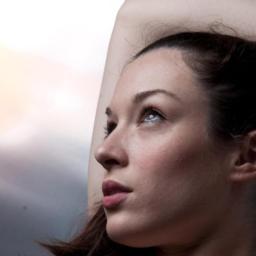} &
\interpfigk{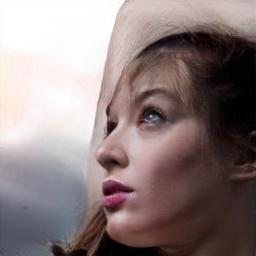} & 
\interpfigk{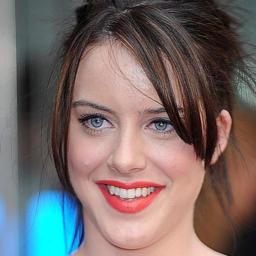} & 
\interpfigk{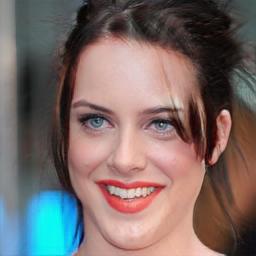} &
\interpfigk{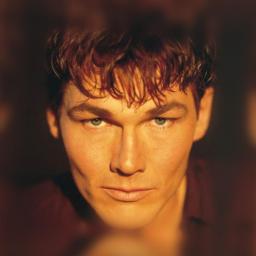} &
\interpfigk{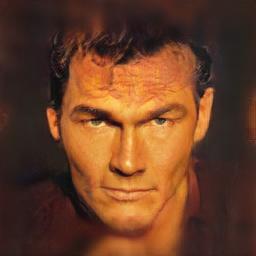} &
\interpfigk{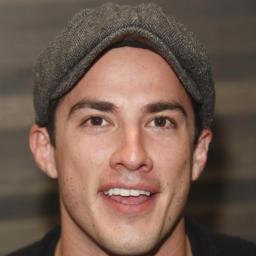} &
\interpfigk{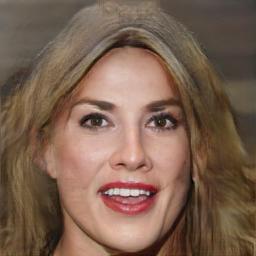} &
\\
Input & Bangs (+) & Input & Bangs (-) & Input & Bangs (-)  & Input & Female (+) \\

\end{tabular}
}
\caption{Failiure cases of VecGAN++.}
\label{fig:failure_cases}
\end{figure*}

\subsection{Analysis of Projected Styles}
We explore the behavior of encoded scales from reference images, $\alpha_t$.
These scales are supposed to provide information about the attribute of the image  (whether a person smiles or not) and its intensity (how big the smile is).
We plot the histograms of $\alpha_t$ values from validation images for smile tags and use orange and blue colors depending on their ground-truth tags from the validation list as shown in Fig. \ref{fig:histog_smileV} and  Fig. \ref{fig:histog_smile} for VecGAN and VecGAN++, respectively.
For the smiling tag, $\alpha_t$ values are mostly disentangled with a small intersection.
For VecGAN, we remove the outliers for visualization purposes. There are some encoded scales far away from the clusters. On the other hand, for VecGAN++ we do not have such a problem and do not remove any data points. 
Other than that, we find VecGAN and VecGAN++ extracted scale attributes to be similar. 
Next, we visualize the samples for the smiling tag for VecGAN++.

Fig. \ref{fig:smile_histo} shows a visualization of validation images plotted based on their $\alpha_t$ values extracted for the smiling tag.
We visualize a few samples from each bin from the histogram above with the same frequency as the histogram value.
The visualization shows that $\alpha_t$ values encode the intensity of the smile.
The rightmost samples have large smiles, and the leftmost samples look almost angry.
On the other hand, the images in the middle space are confusing ones.
We also observe many wrong labeling in the CelebA-HQ dataset by going through the middle space.

We repeat the same analysis for the hair color tag as provided in Fig. \ref{fig:histog_hairV} and  Fig. \ref{fig:histog_hair} for VecGAN and VecGAN++, respectively. Hair color tag is a challenging one as it is expected to have a continuous scale with no clear separation between classes.
We also note that the classifier we train to predict the hair color has an accuracy of $86.6\%$ whereas this score is $94.0\%$ for smile. These scores also show that hair color is more difficult to separate.
We observe that VecGAN struggles to separate attributes for hair color tag, whereas VecGAN++ does a better separation even though it may not be perfect.
We also visualize the validation images  based on their $\alpha_t$ values extracted for hair color tag in Fig. \ref{fig:hair_histo} from VecGAN++.
The images go from black hair to brown hair to blonde hair.
We observe that the shade of hair goes lighter, but we also note that the extracted scales are not perfect, and there is room for improvement. We report the other histogram plots for bangs, eyeglasses, and gender attributes for the VecGAN++ in the last row.
These plots are similar for VecGAN with the exception that VecGAN contains outliers as few samples fall far away from  the distribution and VecGAN++ does not have such problem. 
We omit the plots of VecGAN for these attributes for brevity.

\subsection{Limitations}

In this section, we present the failure cases of our algorithm.
We find our model to struggle with edits when the face are present  with poses that are not very common in the dataset. For example, as shown in Fig. \ref{fig:failure_cases}, in the first row examples and the first example from the second row, faces are rotated and tilted and the model does the edits poorly in these images.
We observe in some edits, artefacts may exist such as in the bangs removal, sometimes they are not completely erased.
Based on our studies and our inspection on other state-of-the-art face editing methods we present in this paper, we conclude that even though great improvements are accomplished, face editing remains an open problem.

\section{Conclusion}

This paper introduces VecGAN++, an image-to-image translation framework with interpretable latent directions.
This framework includes a deep encoder and decoder architecture with latent space manipulation in between.
Latent space manipulation is designed as vector arithmetic where for each attribute, a linear direction is learned.
This design is encouraged by the finding that well-trained generative models organize their latent space as disentangled representations with meaningful directions in a completely unsupervised way.
Therefore, we also extensively compare our method with StyleGAN inversion-based methods and point out their advantages and disadvantages compared to our method.
Each change in the architecture and loss function is extensively studied and compared with state-of-the-arts. Experiments show the effectiveness of our framework.

\ifCLASSOPTIONcaptionsoff
  \newpage
\fi

{
\bibliographystyle{ieee}
\bibliography{egbib}
}

\end{document}